\documentclass[10pt,twocolumn,letterpaper]{article}

\usepackage{3dv}
\usepackage{times}
\usepackage{epsfig}
\usepackage{graphicx}
\usepackage{amsmath}
\usepackage{amssymb}

\usepackage[ruled,vlined]{algorithm2e}
\SetAlFnt{\small}
\usepackage{siunitx}
\usepackage{booktabs}
\usepackage{makecell}
\usepackage{enumitem} %
\usepackage{adjustbox}
\usepackage{subfig}
\usepackage{pifont}%
\newcommand{\cmark}{\ding{51}}%
\newcommand{\xmark}{\ding{55}}%
\usepackage{array}
\newcolumntype{P}[1]{>{\centering\arraybackslash}p{#1}}
\newcolumntype{M}[1]{>{\centering\arraybackslash}m{#1}}

\usepackage{cite}
\usepackage{url}
\usepackage{color}
\newcommand{\mm}[1]{{\color{black}#1}}

\newcommand{\upnew}[1]{{\color{black}#1}}

\threedvfinalcopy %
\def\threedvPaperID{248} %

\newif\ifshowpagenumbers
\showpagenumberstrue

\newcommand{\Sec}{Section~}
\newcommand{\Fig}{Fig.~}
\newcommand{\Tab}{Table~}

\newcommand{\Eq}{Equation.~}

\def\MYTITLE{Event Guided Depth Sensing
}
\title{\MYTITLE}

\newcommand\MYhyperrefoptions{bookmarks=true,bookmarksnumbered=true,
pdfpagemode={UseOutlines},plainpages=false,pdfpagelabels=true,
colorlinks=true,breaklinks=true,
pdftitle={\MYTITLE},%
pdfsubject={Robotics, Computer Vision},%
pdfauthor={M. Muglikar, D. Moeys, D. Scaramuzza},%
pdfkeywords={Depth sensing, Event camera, Asynchronous sensor, Adaptive}}%
\usepackage[\MYhyperrefoptions,pdftex]{hyperref}

\author{Manasi Muglikar \textsuperscript{1}\\
\and
Diederik Paul Moeys \textsuperscript{2}\\
\and
Davide Scaramuzza \textsuperscript{1}\\
}
\newcommand\nomarkerfootnote[1]{%
  \begingroup
  \renewcommand\thefootnote{}\footnote{#1}%
  \addtocounter{footnote}{-1}%
  \endgroup
}
\definecolor{somegray}{rgb}{0.5, 0.5, 0.5}
\newcommand{\darkgrayed}[1]{\textcolor{somegray}{#1}}
\makeatletter
\newcommand*\titleheader[1]{\gdef\@titleheader{#1}}
\AtBeginDocument{%
  \let\st@red@title\@title
  \def\@title{%
    \vskip-4em
    \bgroup\normalfont\large\centering\@titleheader\par\egroup
    \vskip1.5em\st@red@title}
}
\makeatother

\titleheader{\darkgrayed{This paper has been accepted for publication at the\\
International Conference on 3D Vision (3DV), Online, 2021.
\copyright IEEE}}

\ifthreedvfinal
\ifshowpagenumbers
\else
\fi
\fi

\begin{document}

\maketitle

\nomarkerfootnote{\textsuperscript{1}Dept.~Informatics, University of Zurich and Dept.~Neuroinformatics, University of Zurich and ETH Zurich, Switzerland. \textsuperscript{2}Advanced Sensors and Modelling  Group, SONY R\&D Center Europe, SL1.
This research was supported by SONY R\&D Center Europe and the National Centre of Competence in Research (NCCR) Robotics, through the Swiss National Science Foundation.
}
\ifshowpagenumbers
\else
\thispagestyle{empty}
\fi

\global\long\def\bx{\mathbf{x}}
\global\long\def\TE{T_{pc}} %

\global\long\def\taup{\tau_{p}}
\global\long\def\tauc{\tau_{c}}
\global\long\def\bxp{\bx_{p}}
\global\long\def\bxc{\bx_{c}}

\begin{abstract}
Active depth sensors like structured light, lidar, and time-of-flight systems sample the depth of the entire scene uniformly at a fixed scan rate. 
This leads to limited spatio-temporal resolution where redundant static information is over-sampled and precious motion information might be under-sampled.
In this paper, we present an efficient bio-inspired event-camera-driven depth estimation algorithm. 
In our approach, we dynamically illuminate areas of interest densely, depending on the scene activity detected by the event camera, and sparsely illuminate areas in the field of view with no motion.
The depth estimation is achieved by an event-based structured light system consisting of a laser point projector coupled with a second event-based sensor tuned to detect the reflection of the laser from the scene. 
We show the feasibility of our approach in a simulated autonomous driving scenario and real indoor sequences using our prototype.
We show that, in natural scenes like autonomous driving and indoor environments, moving edges correspond to less than 10\% of the scene on average. 
Thus our setup requires the sensor to scan only 10\% of the scene, which could lead to almost 90\% less power consumption by the illumination source.
While we present the evaluation and proof-of-concept for an event-based structured-light system, the ideas presented here are applicable for a wide range of depth sensing modalities like LIDAR, time-of-flight, and standard stereo.
Video is available at \url{https://youtu.be/Rvv9IQLYjCQ}.

\end{abstract}
\section{Introduction}
\label{sec:intro}
\global\long\def\figHeight{4.5cm}
\begin{figure}
    \centering
    \includegraphics[width=\linewidth]{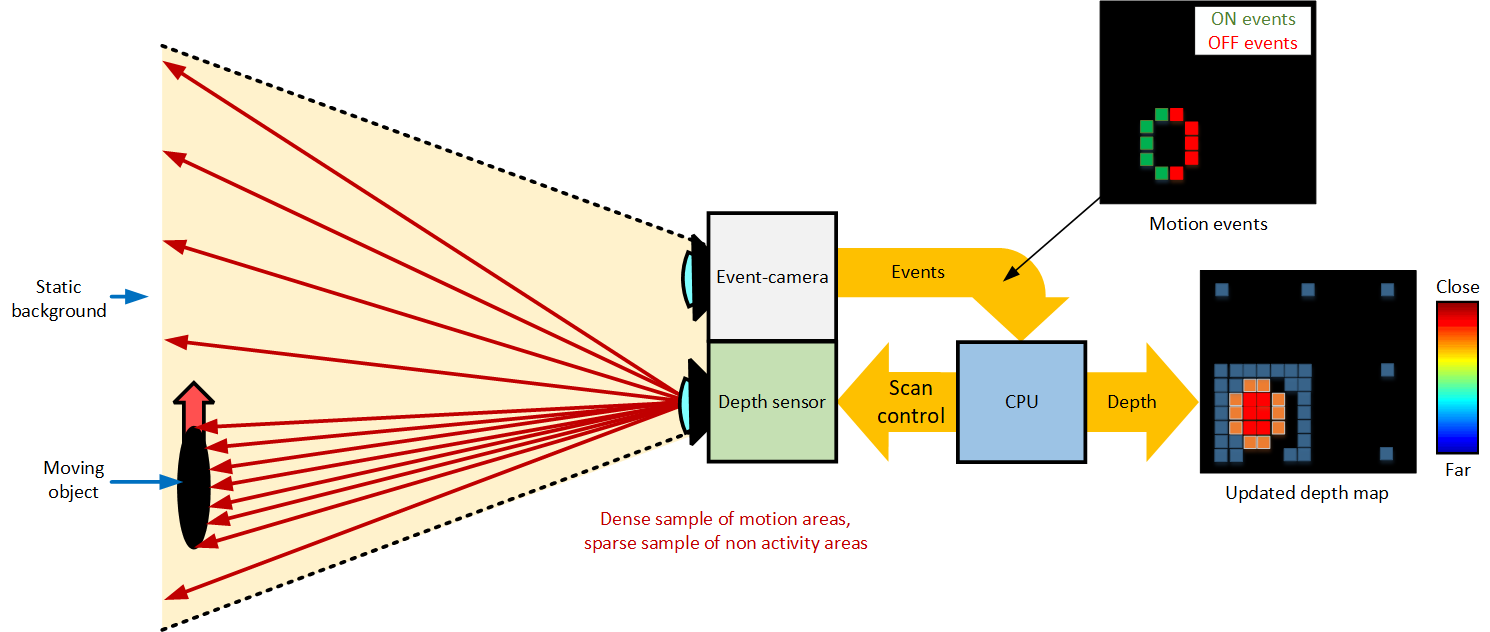}
    \caption{\mm{System overview. 
    Events are triggered by moving objects in the scene or by relative motion.
    These events guide the depth sensor to scan with higher spatial density the affected motion areas of the field of view, resulting in denser depth regions.
    In the rest of the field of view, the depth sensor scans at a lower spatial density, resulting in sparser depth regions.
    }}
    \label{fig:ssystem}
    \vspace{-2ex}
\end{figure}
High-resolution and high-speed depth estimation is desirable in many applications such as robotics, AR/VR, and 3D modeling.
Active depth sensors such as structured light, light detection and ranging (LiDAR), and time-of-flight (ToF) scan the entire scene uniformly at fixed scan rates.
This  leads  to  limited  spatio-temporal resolution where redundant static information is over-sampled  and  precious  motion  information  might  be under-sampled.
We take inspiration from human perception, which involves foveating: areas of interest are scanned with higher spatial resolution while the other regions are scanned with lower spatial resolution.
The area of interest for robotics applications is often characterized by high-contrast features or dynamic objects in the scene; therefore, identifying these areas with low-latency in challenging environments is essential.
We propose to use an event camera to identify these areas of interest as events naturally correspond to moving objects (assuming brightness constancy) and, therefore, do not require further processing to segment the area of interest.

\upnew{Our proposed system is illustrated in \Fig \ref{fig:ssystem}.
The scene consists of a static background with an object moving in front of it.
The motion of the scene triggers events in the event camera corresponding to the edges of the moving object.
These events, represented as an event-frame, are used to guide the depth sensor to scan with higher spatial density the affected motion areas, leading to denser depth samples for the moving object.
The rest of the field of view is scanned sparsely, leading to sparser depth samples for the background.
In scenarios where the setup also undergoes ego-motion (e.g., driving), events would correspond to edges of the scene, triggered by moving objects or relative motion.
Thus, the depth samples would have higher resolution in the neighbourhood of scene edges and sparse resolution in constant-intensity regions of the scene.
}

As a depth-sensing modality, in this work we consider event-based structured light (SL).
\upnew{Event-based SL systems provide high-speed depth estimation compared to their frame-based counterparts \cite{Brandli13fns, Matsuda15iccp, Martel18iscas, muglikar213dv} due to the high temporal resolution of event cameras.
However, their accuracy also drops at higher spatial resolutions due to noise in event data.}
While we present the evaluation and proof-of-concept for an event-based SL system, the ideas presented here are applicable for a wide range of depth sensing modalities like LIDAR, ToF, and standard stereo.
Note that we do not want to replace full-frame depth sensors, but rather provide a complementary depth sensor that can scan dynamic objects with low-latency.
We summarize our \textbf{contributions} below: 
\begin{itemize}
    \item A novel formulation of event-guided foveating depth sensing to guide a depth sensor to scan moving objects with higher spatial resolution and low-latency.
    \item Real-world experimental evaluation on dataset recorded with a prototype.
    \item Neural-network based depth completion from event-guided sparse depth samples.
\end{itemize}

The paper is structured as follows: In \Sec \ref{sec:related-work:eventSL}, we give an overview of event-based structured light systems. 
\Sec \ref{sec:method} presents the problems with high resolution and high frequency event-based SL systems and non-idealities of event cameras. 
\Sec \ref{sec:method:ours} details our proposed solution for event-guided depth estimation.
In \Sec \ref{sec:method:depthcomplete} we provide details for depth completion from sparse event-guided depth.
Finally, we evaluate our method in simulation and real world in \Sec \ref{sec:results} and conclude in \Sec \ref{sec:conclusion}.
\section{Related works}
\label{sec:related-work}
\begin{table}
    \centering
    \begin{adjustbox}{max width=\columnwidth}
    \setlength{\tabcolsep}{3pt}
    \begin{tabular}{|l|l|l|l|l|}
    \toprule
    \textbf{Method}  &  \textbf{Depth sensor} &  \textbf{Modality} & \textbf{Frame rate} & \textbf{Adaptive} \\ \midrule
    Intel RealSense & SL (stereo) & I & 90 & \color{red}\xmark \\
    Azure Kinect & ToF & I & 30 & \color{red}\xmark \\
    Programmable light curtains \cite{Wang18eccv} & SL & I & 5.6 & \color{green}\cmark \\
    Pittaluga et al. \cite{Pittaluga203dv} & ToF & I & 24 & \color{green}\cmark \\ \midrule
    Brandli et. al. \cite{Brandli13fns} & SL & E & - & \color{red}\xmark \\
    Matsuda et. al. \cite{Matsuda15iccp}  & SL & E & 60 & \color{red}\xmark\\
    Martel et. al. \cite{Martel18iscas} & SL(stereo) & E & - & \color{red}\xmark\\
    Muglikar et. al. \cite{muglikar213dv} & SL & E & 60 & \color{red}\xmark\\
    \textbf{Ours} & SL & E & 60 & \color{green}\cmark\\
    \bottomrule
    \end{tabular}
    \end{adjustbox}
    \vspace{-1ex}
    \caption{\label{tab:related-work}Summary of active depth sensors. 
    The modality indicates the sensor used in the setup for depth estimation (\textit{I} is for standard camera images and \textit{E} stands for event camera).}
    \vspace{-2ex}
\end{table}

Prior work on active depth sensors is summarized in the \Tab \ref{tab:related-work}.
The first half focuses on frame-based active depth sensors (modality \textit{I}) and the second half focuses on event-based active depth sensors (modality \textit{E}).
We first discuss literature on event-based active depth in \Sec \ref{sec:related-work:eventSL} followed by a review of adaptive sampling depth sensors (frame-based SL or ToF) in \Sec \ref{sec:related-work:adaptiveSL}.
Lastly, we summarize some of the other sensing modalities that make use of foveation for different applications in \Sec \ref{sec:related-work:adaptivesensors}.
\subsection{Event-based active depth}
\label{sec:related-work:eventSL}
Since event cameras are novel sensors (commercially available since $2008$), the literature on event-based structured light systems is not vast.
\Tab \ref{tab:related-work} summarizes some of the event-based SL systems.
The earliest work \cite{Brandli13fns} combined a DVS with a pulsed line laser to reconstruct a terrain.
The pulsed laser line was projected at a fixed angle with respect to the DVS while the terrain moved beneath it.
In \cite{Martel18iscas}, they combined a laser light source with a \emph{stereo} setup consisting of two DAVIS240 cameras~\cite{Brandli14ssc}.
The laser illuminated the scene and the synchronized event cameras recorded the events generated by the reflection from the scene.
Hence the light source was used to generate stereo point correspondences, which were then triangulated (back-projected) to obtain a 3D reconstruction.
The SL system MC3D \cite{Matsuda15iccp} comprised of a laser point projector and a DVS. 
The laser scanner moved in a raster pattern across the scene.
The DVS captured the reflection of the light from the scene as events.
MC3D then converted temporal information of events at each pixel into disparity.
It exploited the redundancy suppression and high temporal resolution of the DVS, also showing appealing results in dynamic scenes compared to frame-based counterparts.
\upnew{Previous systems demonstrated the capabilities of event camera for SL with a low-resolution event sensor.
Recently, \cite{muglikar213dv} used a high resolution event camera for SL and discuss the noises present in event timestamps (jitter and BurstAER effects).
They proposed a depth estimation algorithm to improve the robustness of event-based SL up to a certain extent.
However, as higher resolution event cameras are emerging, the accuracy of event-based SL systems does not scale with the resolution as noise in the event timestamps increases significantly due to the high event rate.} %

\subsection{Adaptive depth}
\label{sec:related-work:adaptiveSL}
One of the earliest works used programmable light curtains to detect objects at fixed shapes around the robot or car \cite{Wang18eccv}.
Recently, \cite{Pittaluga203dv} showed a proof of concept prototype with camera-based foveating LIDAR showing the advantage of the motion-based adaptive sampling.
However, they were limited by the hardware and camera to $24Hz$ depth acquisition time.
\subsection{Adaptive foveating sensors}
\label{sec:related-work:adaptivesensors}
The concept of attention-driven sensing is inspired by biology. It follows the idea that the human eye more efficiently concentrates its processing in areas of high interest (high contrast or motion) while under-sampling redundant information (static part of the background or low contrast features). In the retina, the attention area is focused in the fovea, which is densely populated with photoreceptors and retinal ganglion cells  \cite{Remington3ed}. This approach has found interest in computer vision for various algorithms including 3D SLAM \cite{Charrow15, Cadena16tro}.  To avoid rendering entire frames, foveation has also been used in AR/VR headsets \cite{Guenter12siggraph}

\section{Event-based SL}
\label{sec:method}
In this section, we introduce the basics of event-camera-projector setup.
We discuss the sources of noise (\Sec \ref{sec:method:noise}) and evaluate the theoretical performance of this system at high resolutions and projector scanning frequencies (\Sec \ref{sec:method:highres}).
\subsection{Basics of event-based SL}
\label{sec:method:basics}
We consider the problem of depth estimation using a laser point projector and event camera.
The scene is illuminated by the laser beam inside the projector, moving in a raster scanning fashion.
This light reflected from the scene generates events\footnote{An event camera generates an event $e_k = (\bx_k,t_k,p_k)$ at time $t_k$ when the logarithmic brightness at the pixel $\bx_k=(x_k,y_k)^\top$ increases or decreases by predefined threshold $C$:
$L(\bx_k,t_k) - L(\bx_k,t_k-\Delta t_k) = p_k C$, where $p_k \in \{-1,+1\}$ is the sign (polarity) of the brightness change, and $\Delta t_k$ is the time since the last event at the same pixel location.} in the event camera.
Assuming the travel time for light from the projector to the camera is small, we can map the event pixel to the corresponding projector pixel using the event timestamp.
Depth is estimated by triangulating the pixel from the event-camera plane and the projector plane \cite{muglikar213dv}.
Depth, thus, depends on accurate event timestamps.
We now take a look at the non-idealities of events that lead to noise in the timestamps.

\subsubsection{Event camera imprecision sources} 
\label{sec:method:noise}
\begin{figure}[t]
    \centering
    \includegraphics[width=\linewidth]{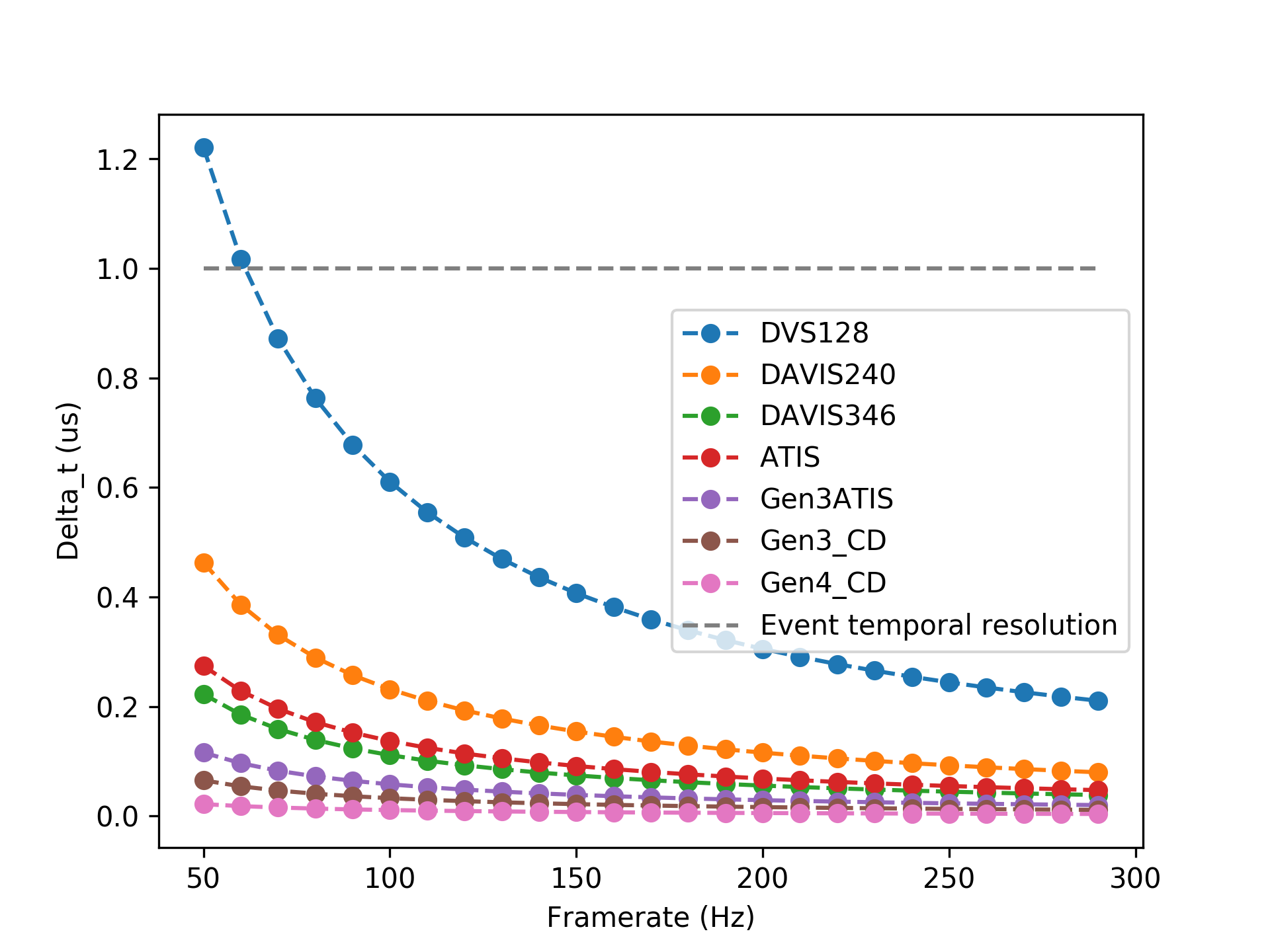}
    \caption{\mm{Theoretical} effect of projector frame rate and resolution on time between consecutive events. We observe that with an increase in the framerate from \SI{50}{Hz} to \SI{100}{Hz}, for the DVS128 (lowest resolution) sensor, the time between two consecutive events drops below the event temporal threshold of \SI{1}{\micro \second} }
    \label{fig:delta_t_vs_fps}
\end{figure}

Available event cameras have a minimum possible temporal resolution of \SI{1}{\micro \second}.
However, in practice, latency, transistor noise and variable readout delays introduce noise in the data, which decrease the precision of the event timestamp. Such non-idealities are design and manufacturer specific.
\textit{Latency}: defined as the time it takes for an event to be registered since the moment logarithmic change in intensity exceeds the threshold, it can currently range on average from a few microseconds to hundreds of milliseconds, depending on bias settings, manufacturing process and illumination level. %
\textit{Transistor noise}: defined as the random transistor noise of the circuits, it also depends on settings and illumination. %
This noise randomly changes the measured signal, leading to threshold-comparison jitter \cite{Moeys17tbcas}.
\textit{Other non-idealities}: encompassing parasitic photocurrents and junction leakages, these effects bias event generation to one specific polarity \cite{Nozaki17ted}.
\textit{Read-out architecture}: arbitrated architectures, which preserve or partially preserve the order of the pixels' firing, lead to significant queuing delays before the timestamping operation. This is particularly noticeable when the number of active pixels (and resolution) scale up. Scanning readouts, on the other hand, limit the possible delays by sacrificing event timing resolution.
\textit{Jitter}: defined as the random variation that appears in timestamps, it depends on all of the aforementioned factors, all of which increase the unpredictability and imprecision of the event timing.
\subsubsection{Performance at high resolution and scanning frequencies}
\label{sec:method:highres}
Accuracy of event-based SL system depends on the accuracy of event timestamps.
Therefore, for consecutive events triggered by the laser beam to be uniquely distinguished, the timestamps should differ by at least \SI{1}{\micro \second}, which is the event camera timestamp resolution.
The time it takes for laser beam of the projector to move to the next pixel, $\Delta t$, is inversely proportional to the scanning frequency $f$ and spatial resolution $W \times H$ pixels:
\begin{equation}
\Delta t = \frac{1}{f \times W \times H}
\end{equation}
We define this approach as dense sampling since it illuminates every pixel.
\upnew{We consider commercially available event cameras covering a wide range of resolutions namely: DVS128 (resolution $128 \times 128$), DAVIS240 (resolution  $240 \times 180$), DAVIS346 (resolution $346 \times 260$), Prophesee ATIS (resolution $302 \times 240$), Prophesee Gen3\_CD (resolution $640 \times 480$), Prophesee Gen3ATIS (resolution $480 \times 360$), Prophesee Gen4\_CD (resolution $1280 \times 720$), and compute the theoretical $\Delta t$ for each sensor while varying the projector framerate from \SI{50}{Hz} to \SI{290}{Hz}.
Projector framerate here refers to the scanning frequency of the laser point projector.}
As we see in ~\Fig \ref{fig:delta_t_vs_fps}, $\Delta t$ is lower than $\SI{1}{\micro\second}$ for almost every sensor resolution for scanning frequency higher than $\SI{60}{Hz}$.
Since $\Delta t$ depends on the scanning frequency and sensor resolution, there is no simple way to ensure $\Delta t$ is higher than event camera temporal resolution.

\begin{figure}[t]
    \centering
    \includegraphics[width=\linewidth]{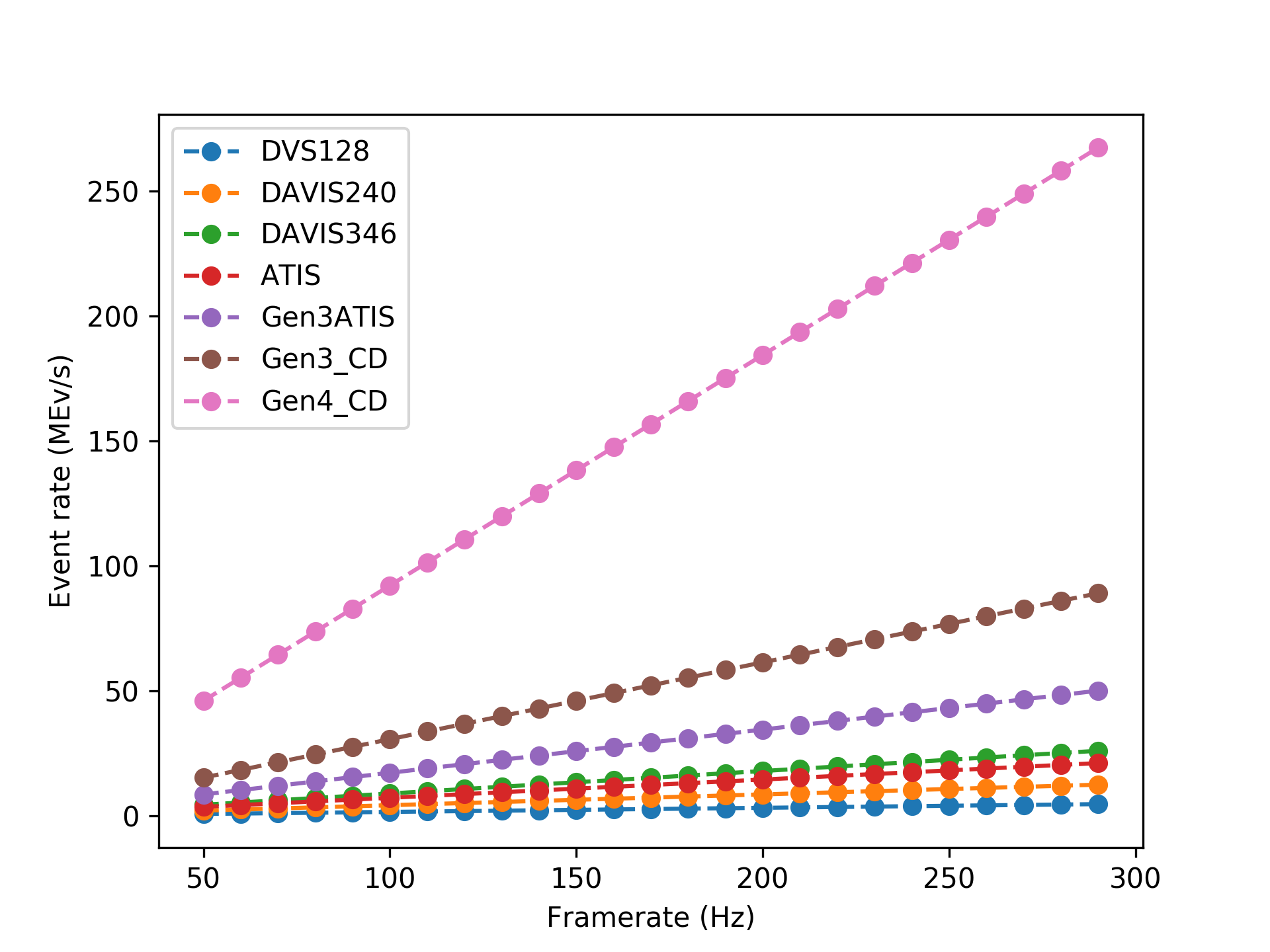}
    \caption{\mm{Theoretical} effect of projector frame rate and resolution on event-rate of an event-based structured light system.}
    \label{fig:event_rate}
\end{figure}

Similarly, another disadvantage of dense sampling arises from the high-event rates.
In event-based SL, the event-rate (measured in MEv/s) is directly proportional to the scanning frequency and spatial resolution.
Theoretical event-rates for multiple projector scanning frequencies and event camera resolutions are shown in \Fig \ref{fig:event_rate}.
\upnew{For lower resolution cameras such as DVS128 \cite{Lichtsteiner08ssc}, the increase in the event-rate with respect to the scanning frequency is not significant.
Whereas for a higher resolution camera such as Prophesee Gen4\cite{Finateu20isscc}, the event rate increases from $44$ MEv/s (scanning frequency of $50$ Hz) to $265$ MEv/s (scanning frequency of $290$ Hz).
Since event rate plays a significant role in the event timestamp error\cite{Ryu19cvprw}, the errors in depth will be significant in high-resolution and high-speed event-based structured light systems.
At a high event rate, of $10$ MEv/s, the timestamp error is in the order of \SI{10}{\micro\second} which can go up to \SI{200}{\micro\second} for event rates of $265$ MEv/s.
}
Sparse sampling mitigates some of these errors by reducing the event rate.

\section{Sparse sampling}
\label{sec:method:sparse}
Sparse sampling increases the $\Delta t$ between consecutive events by scanning a fraction of the pixels as shown in \Fig \ref{fig:sparse_viz}.
The $\Delta t$ is now proportional to the number of pixels skipped between two illuminated pixels (defined as $N$) as shown in \Eq \ref{eq:deltatsparse}.
This allows us to control the timestamp errors due to small $\Delta t$ up to a certain extent.
\begin{equation}
\label{eq:deltatsparse}
\Delta t = \frac{N}{f \times W \times H}
\end{equation}

\global\long\def\figHeight{3.9cm}
\begin{figure}[t]
    \centering
    \subfloat[Dense sampling]{\label{fig:dense}\includegraphics[trim={0 5cm 15cm 0cm},clip,height=\figHeight]{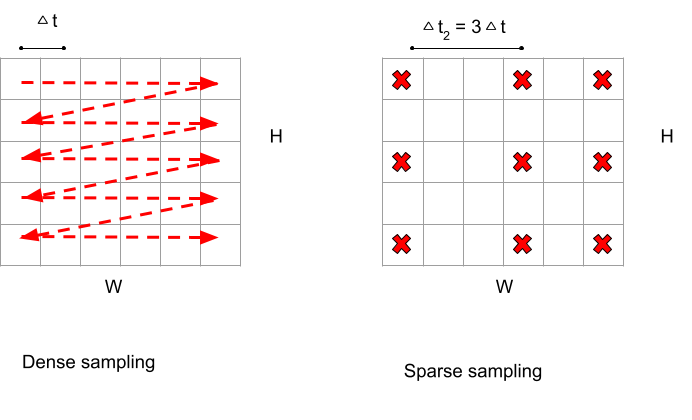}}\;
    \subfloat[Sparse sampling]{\label{fig:sparse}\includegraphics[trim={13cm 5cm 2cm 0cm},clip,height=\figHeight]{images/desne_vs_sparse_sampling.png}}\;\\
    \caption{Visualization of dense sampling versus sparse sampling. The $\Delta t$ between consecutive events for dense sampling  is very small. On the other hand, $\Delta t$ for sparse sampling is $3$ times higher than dense in this example.}
    \label{fig:sparse_viz}
\end{figure}
Sparse sampling, however, does not consider the scene geometry.
This can lead to undersampling of areas of fast moving objects and oversampling planar areas, leading to redundant information.
Conversely, sparse sampling is not always desirable in areas of interest (usually of object motion), as seen in \Fig \ref{fig:real_events:no_moving}.
In an autonomous driving scenario, for example, if a fast moving vehicle or pedestrian moves in front of a car, it is useful to have more depth information about this. 
For robotics applications, dynamic objects in the scene are important for obstacle avoidance or local planning of trajectories.
These scenarios make it desirable to have a higher resolution in these areas of interest at a low-latency.

\section{Event-guided depth sensing}
\label{sec:method:ours}
Sparse sampling has the disadvantage of under sampling areas of interest and over sampling planar areas.
We therefore propose to use an event-based sampling for structured light depth with event cameras.
As events naturally encode the relative motion of the scene with high temporal resolution, they are ideal for this task.
Our approach, is shown in \Fig \ref{fig:ssystem}.
The motion of the scene triggers events in the event camera corresponding to the edges of the moving object.
Assuming brightness constancy, events observed by the event camera correspond to edges moving either due to ego-motion or object motion.
The distinction between these two types of events (\ie ego-motion events and object motion events) is simple in the case of a stationary camera since all the events are caused due to object motion.
On the other hand, differentiating these two motion types is not simple in the case of a moving setup and is typically solved using optimization \cite{Stoffregen19iccv} or by assuming only rotational motion using IMU priors \cite{Falanga19ral}.
Since our application aims for low-latency depth sensing, we consider all the events (ego-motion and object motion) as equal contributors and do not distinguish between them.
Future work could embed fined tuned faster algorithms for ego motion segmentation onto ASICs or FPGAs in order to reduce latency overhead. 
These events, represented as an event-frame, are used to guide the depth sensor to scan with higher spatial density the affected motion areas, leading to denser depth samples for the moving object.
The event camera is aligned with the depth sensor via rectification.
The rest of the field of view is scanned sparsely, leading to sparser depth samples for the background.
In practice, we apply a median-filter on the event-frame to reduce the influence of noise.
On this event-frame, we use simple contour detection to get a bounding box around events.
\Fig \ref{fig:real_events:moving} shows the sampling strategy described here.
Specifically for event-based structured light system, the depth sensor consists of a laser beam projector and a second event camera tuned to observe the events generated by the illumination source.
Assuming the event camera is synchronized with the projector, events due to the projector in one scan period are represented as a time surface \cite{Gallego20pami}(example time surface shown in \Fig \ref{fig:real_events}).

\global\long\def\figHeight{3.9cm}
\begin{figure}[t]
    \centering
    \subfloat[Sparse sampling]{\label{fig:real_events:no_moving}\includegraphics[trim={5cm 2cm 6cm 4cm},clip,height=\figHeight]{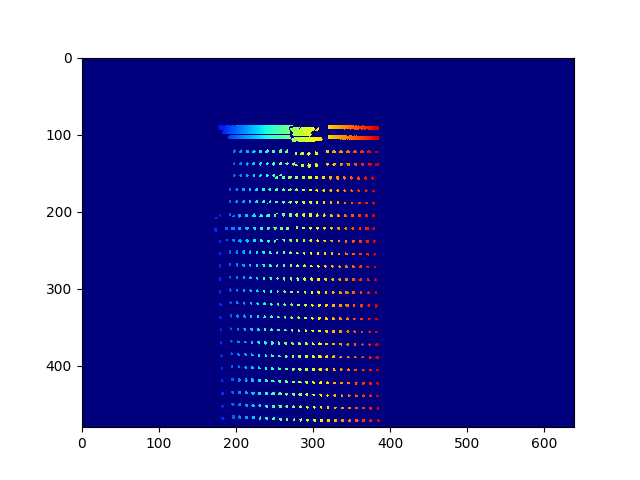}}\;
    \subfloat[Event-guided sampling]{\label{fig:real_events:moving}\includegraphics[trim={5cm 2cm 6cm 4cm},clip,height=\figHeight]{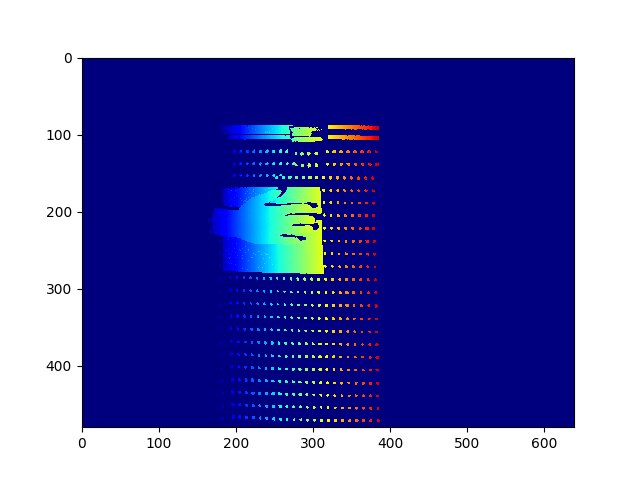}\;\;\includegraphics[height=\figHeight]{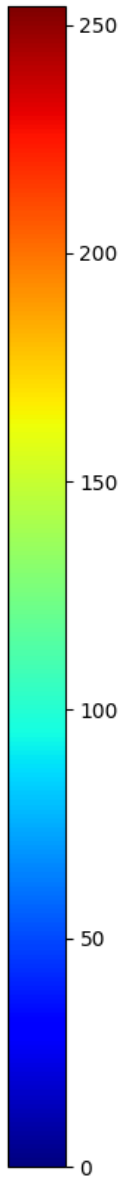}}\\
    \caption{Example event time map with sparse sampling and event-guided sampling.}
    \label{fig:real_events}
\end{figure}

\subsection{Depth completion with event-based sampling}
\label{sec:method:depthcomplete}
While we conservatively sample the scene, a natural question to ask is "can we recover the complete depth profile from sparse depth measurements?"
Several works address the problem of full scene depth estimation from sparse samples \cite{Ma16iros, Uhrig173dv, Eldesokey20cvpr, Cheng18eccv}.
Recently, \cite{Ma17arxiv} formulated the depth completion problem with compressive sensing literature and provided exactness and stability analysis for 3D depth completion.
Proposition $6$ of \cite{Ma17arxiv} showed that exact recovery of depth from sparse set of samples is possible if the sparse sample set covers the edges and their adjacent pixels.

Previous approaches estimated dense depth from monocular events \cite{Hidalgo20threedv} or combination of events and frames \cite{Gehrig21ral}.
In our approach, we supplement the events with depth samples generated with event-guided approach to estimate dense depth.\\
\textbf{Network Architecture} The general network architecture is inspired from U-Net \cite{Gehrig21ral, Hidalgo20threedv}, which was used previously used for monocular depth estimation with events and combining events with images.
Similar to \cite{Gehrig21ral} our architecture uses skip connections and residual block.
However, unlike their approach, we do not combine events with images, but rather with event-guided depth samples.\\
\textbf{Depth representation}: The metric depth $D_m$ is first converted into a normalized log depth map $D \epsilon [0,1]$ facilitating large depth variation learning \cite{Gehrig21ral, Hidalgo20threedv, Li18cvpr, Eigen15iccv}.
Logarithmic depth is computed similar to \cite{Gehrig21ral} as:
\begin{equation}
    D = \frac{1}{\alpha} log \frac{D_m}{D_{max}}+1
\end{equation}
where, $D_{max} = \SI{1000}{ \meter}$ and $\alpha = 5.7$.\\
\textbf{Event representation}: Events are converted to fixed-size tensor to utilize existing convolutional neural network architectures.
Events in a time window $\Delta T$ are draw into a voxel grid with spatial dimensions $H \times W$ and $B$ temporal bins \cite{Zhu19cvpr, Rebecq19pami, Gehrig19iccv}.
\begin{equation}
    V(x,y,t) = \sum p_i \delta(x-x_i, y-y_i) max\{0, 1-|t-t_i^*|\}
\end{equation}
where $t_i^* = \frac{B-1}{\Delta T} (t_i - t_0)$. The number of bins $B$ is set to 5 for all experiments \cite{Rebecq19pami}.
\section{Experiments}
\label{sec:results}
To the best of our knowledge, there is no dataset available for event-based structured light systems to evaluate the proposed method.
We therefore evaluate our method by building the prototype and collecting a custom dataset.
Details about the prototype are presented in \Sec \ref{sec:exp:prototype}.
We also evaluate our method in simulation with EventScape dataset, a new dataset with events, intensity frames, semantic labels, and dense depth maps recorded  in  the  CARLA  simulator\cite{Hidalgo20threedv}.
Since our method is agnostic to depth sensing modality (SL, ToF or LiDAR), we show the advantage of event-guided depth sensing over dense sampling.
Finally, we show the result of depth completion on the simulation dataset.
\subsection{Prototype}
\label{sec:exp:prototype}
\global\long\def\figHeight{4.0cm}
\begin{figure}[t]
    \centering
    \includegraphics[trim={46cm 40cm 15cm 29cm},clip,height=\figHeight]{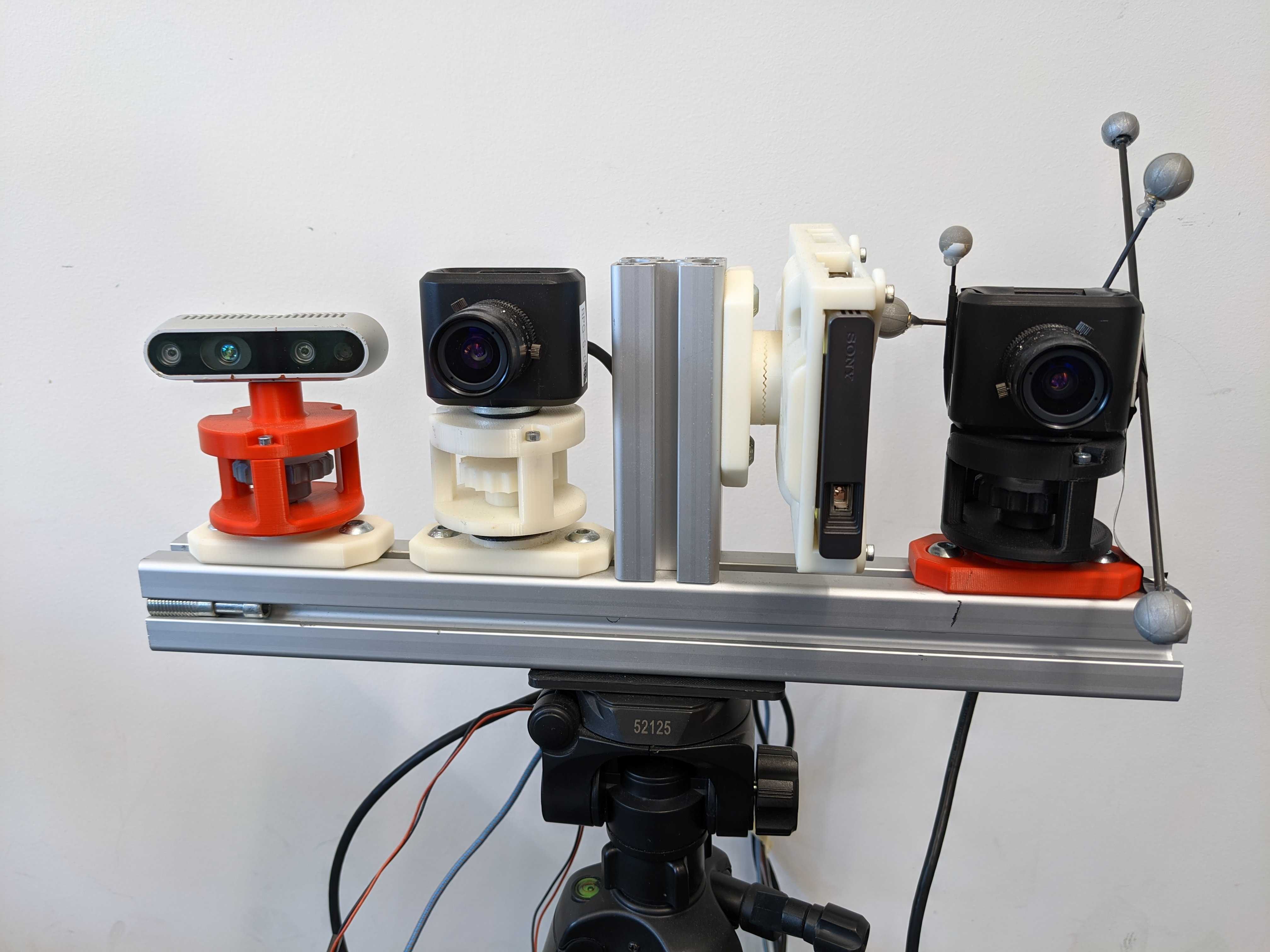}
    \caption{Our prototype for event-based SL system consists of Prophesee Gen3 event camera (mounted on white plate) and Sony MP-CL1A laser point projector. 
    The second event camera (mounted on red plate ) guides the laser projector to illuminate the active areas of the scene.
    This second event camera is next to the projector and aligned via rectification and alignment. }
    \label{fig:setup}
\end{figure}%
Our prototype for event-based structured light shown in \Fig \ref{fig:setup} consists of a Prophesee Gen3 event camera with resolution of $640 \times 480$.
The projector is a laser point projector Sony MP-CL1A with resolution of $1920 \times 1080$ and scanning frequency of $\SI{60}{Hz}$.
The calibration of the event camera uses \cite{Muglikar2021CVPR}.
We use a second Prophesee Gen3 event camera (resolution $640 \times 480$) for guiding the SL depth sensor.
This event camera is separated from the projector with a baseline of \mm{$\SI{4}{\centi \meter}$}.
The event camera is aligned with the projector via rectification and alignment.
For each sequence, alignment is performed using feature-based homography estimation by projecting a checkerboard pattern on the scene and representing the events as an event frame to extract the corner point.
Note that while this alignment would work, it requires the objects to move in a front-parallel plane.
To allow for alignment mismatch, we dilate the pixels in the projector plane to illuminate a larger field of view.
To ensure the second event camera only observes the events and not the events due to projector illumination, we adjust the bias settings such that we filter out high frequency changes primarily caused by the projector.
In particular, we reduce the bandwidth of the source follower buffer, which follows the logarithmic photoreceptor.
\subsection{Evaluation}
We evaluate the proposed approach in simulation and with a prototype.
We compare the area of pixels that is required to scan with our approach in \Sec \ref{sec:exp:activepixels} in both simulation and real world sequences.
We evaluate the performance of our approach against dense and sparse sampling techniques with our prototype in \Sec \ref{sec:exp:recon}.
Lastly, we show the results of our depth completion network in \Sec \ref{sec:exp:depthcompletion}.

\subsubsection{Active pixels} 
\label{sec:exp:activepixels}
Active pixels define the area to be scanned.
For an event-guided sensing, this area is data driven.
If the scene activity is present in all the pixels, this method would be equivalent to dense scanning.
However, in a natural scene, the active area corresponding to moving objects is usually small compared to the sensor's field of view (FoV).
We evaluate the activity of scene on the image plane in two scenarios ($i$) stationary setup and moving objects ($ii$) autonomous driving scenario where the sensor is moving (ego-motion) along with moving objects in the scene.
Since we do not differentiate between events from object motion and events due to ego-motion, we associate all the events to areas with activity.\\
\textbf{Simulation}
For simulation, we use the EventScape dataset which demonstrates the autonomous driving scenario.
Since events arise both due to ego-motion and moving objects, the active pixels depend on the scene.
We calculate active pixels for over $116$ test sequences.
The average active pixels over all the sequences is $7.1\%$ of the resolution (camera resolution of $256 \times 512$)
For the sequences that have significant motion distributed all over the scene such as $T_{064}$, this still amounts to $20\%$ of the total resolution as seen in \Fig \ref{fig:experim:depthcompletion}.
\\
\textbf{Real Dataset} 
We evaluate our method on multiple dynamic scenes \Fig \ref{fig:recons} with diverse motion scenarios to show our system's capabilities. 
The percentage of active pixels in these sequences are in the range of 0.73\% to 10.34\%.
In the stationary scenario, active pixels are minimum (0.73\%) as they correspond to sparse samples.
The maximum active pixels are triggered (10\%) with the moving setup as this creates significant activity in the entire scene due to ego-motion.
By sampling a small area of pixels, this system can lead to a low-power and high framerate depth sensor.

\subsubsection{Reconstruction}
\label{sec:exp:recon}

\global\long\def\figWidthRef{0.18\linewidth}
\global\long\def\figWidth{0.18\linewidth}
\begin{figure}[t]
	\centering
	\setlength{\tabcolsep}{2pt}
	\begin{tabular}{
			M{0.2cm}
			M{\figWidthRef}
			M{\figWidth}
			M{0.2cm}
			M{\figWidthRef}
			M{\figWidth}}
		& Scene & \textbf{Depth}& & Scene & \textbf{Depth}
		\\
		\rotatebox{90}{\makecell{Moving hand}}
		&\includegraphics[trim={0cm 15cm 15cm 10cm},clip,width=\linewidth]{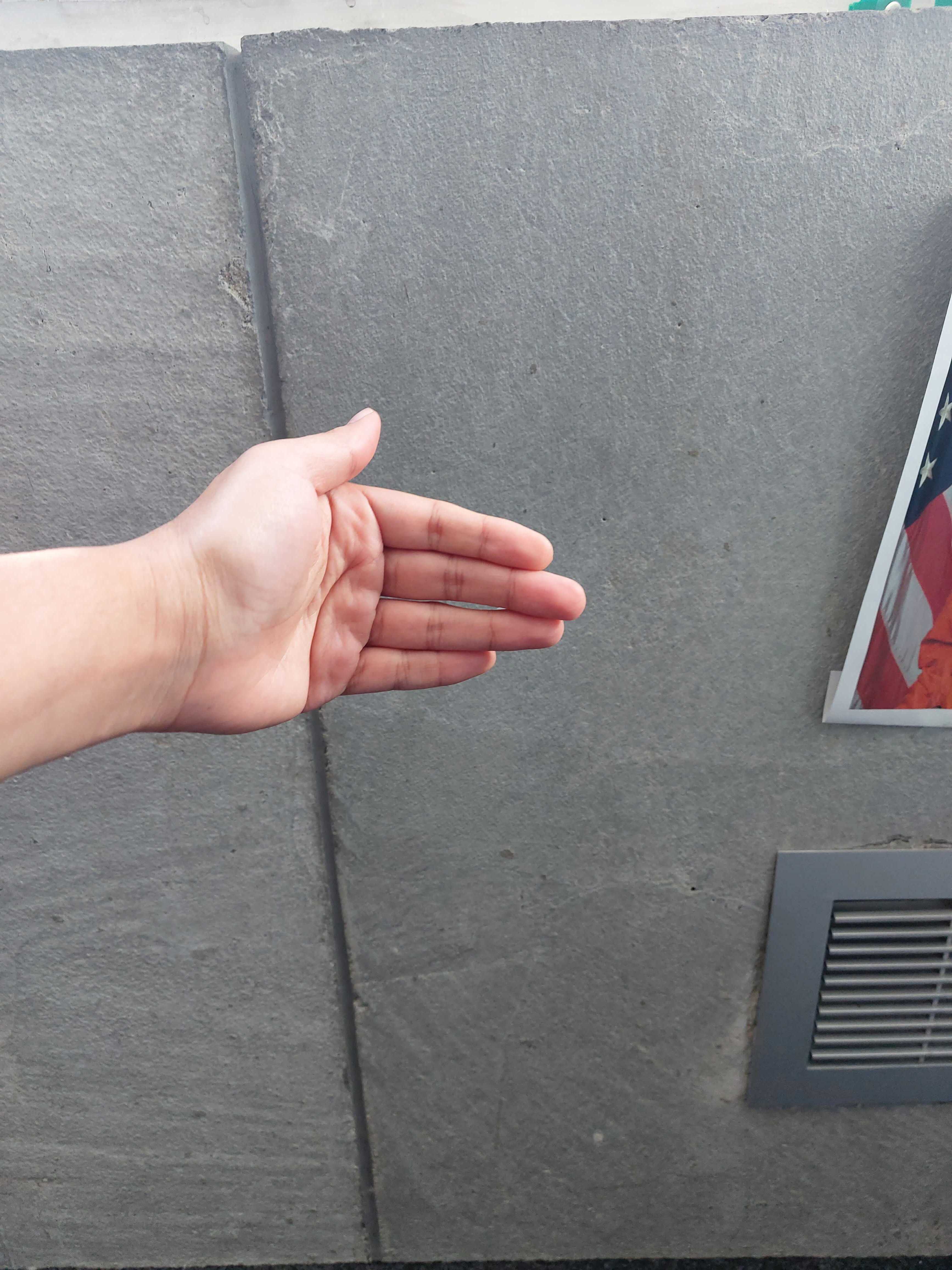}
		&\frame{\includegraphics[trim={15cm 5cm 25cm 8cm},clip,width=\linewidth]{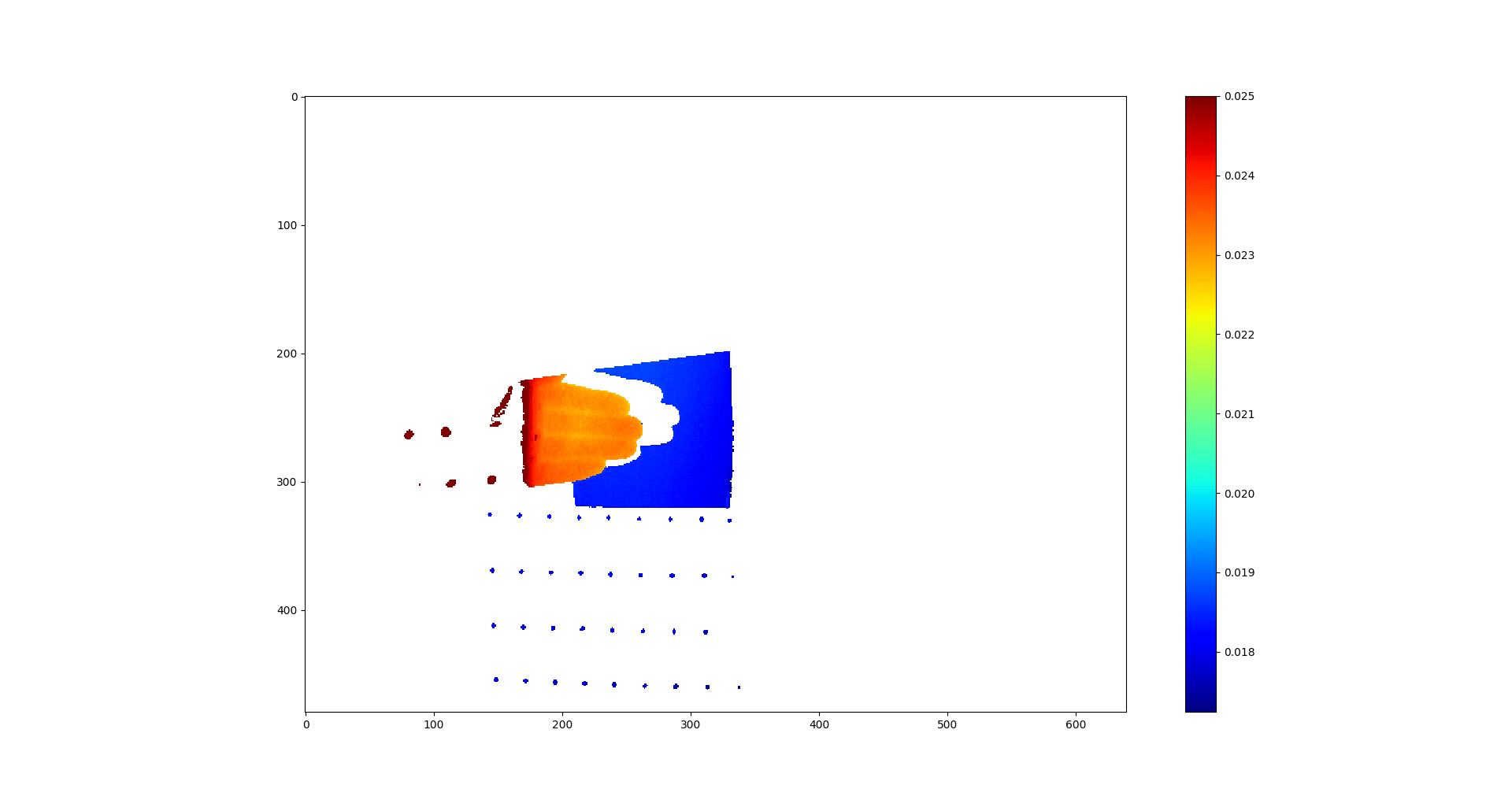}}
		&\rotatebox{90}{\makecell{\mm{Ball}}}
		&\includegraphics[trim={1cm 15cm 0cm 5cm},clip, width=\linewidth]{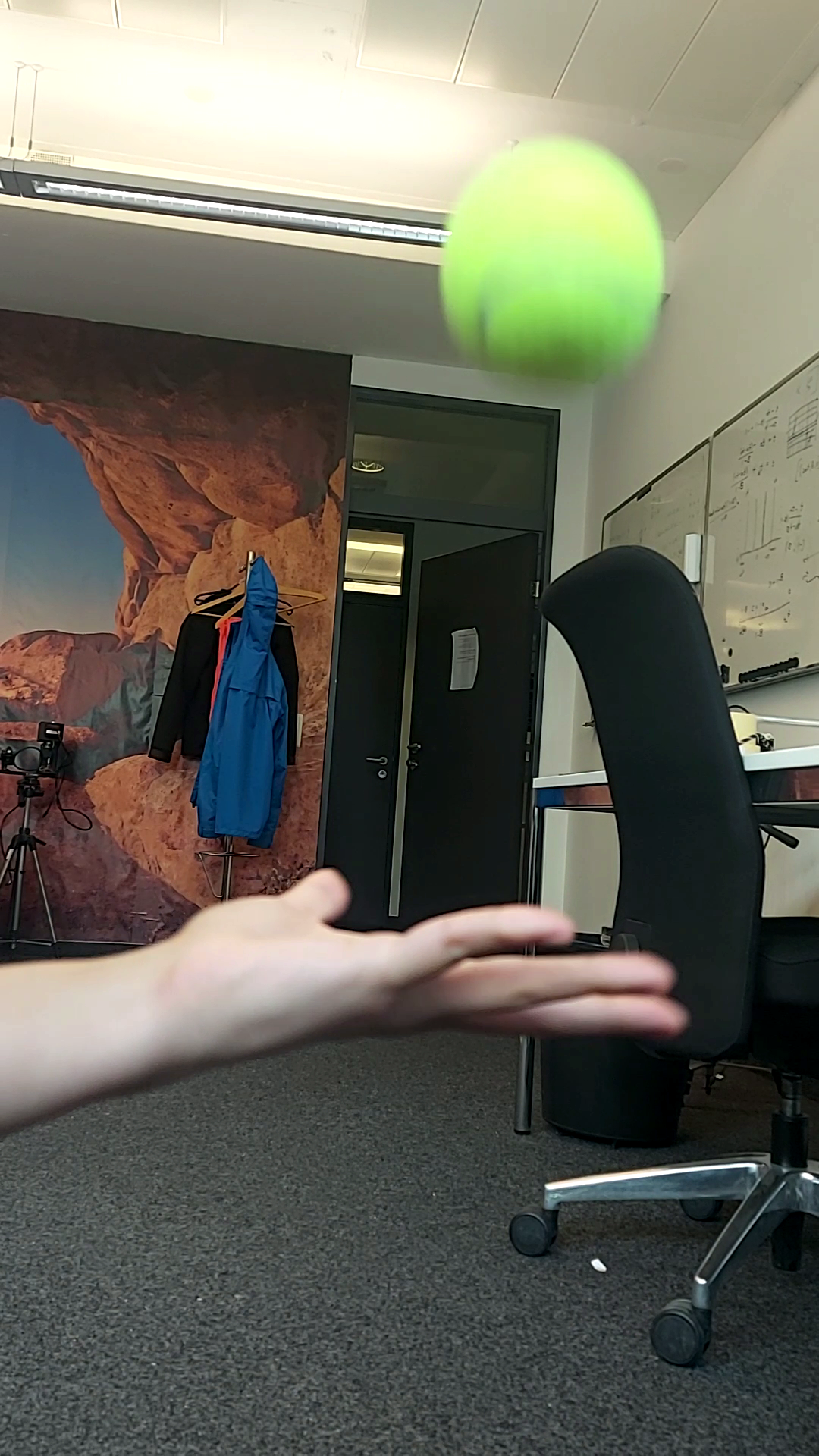}
		&\frame{\includegraphics[trim={15cm 5cm 25cm 8cm},clip,width=\linewidth]{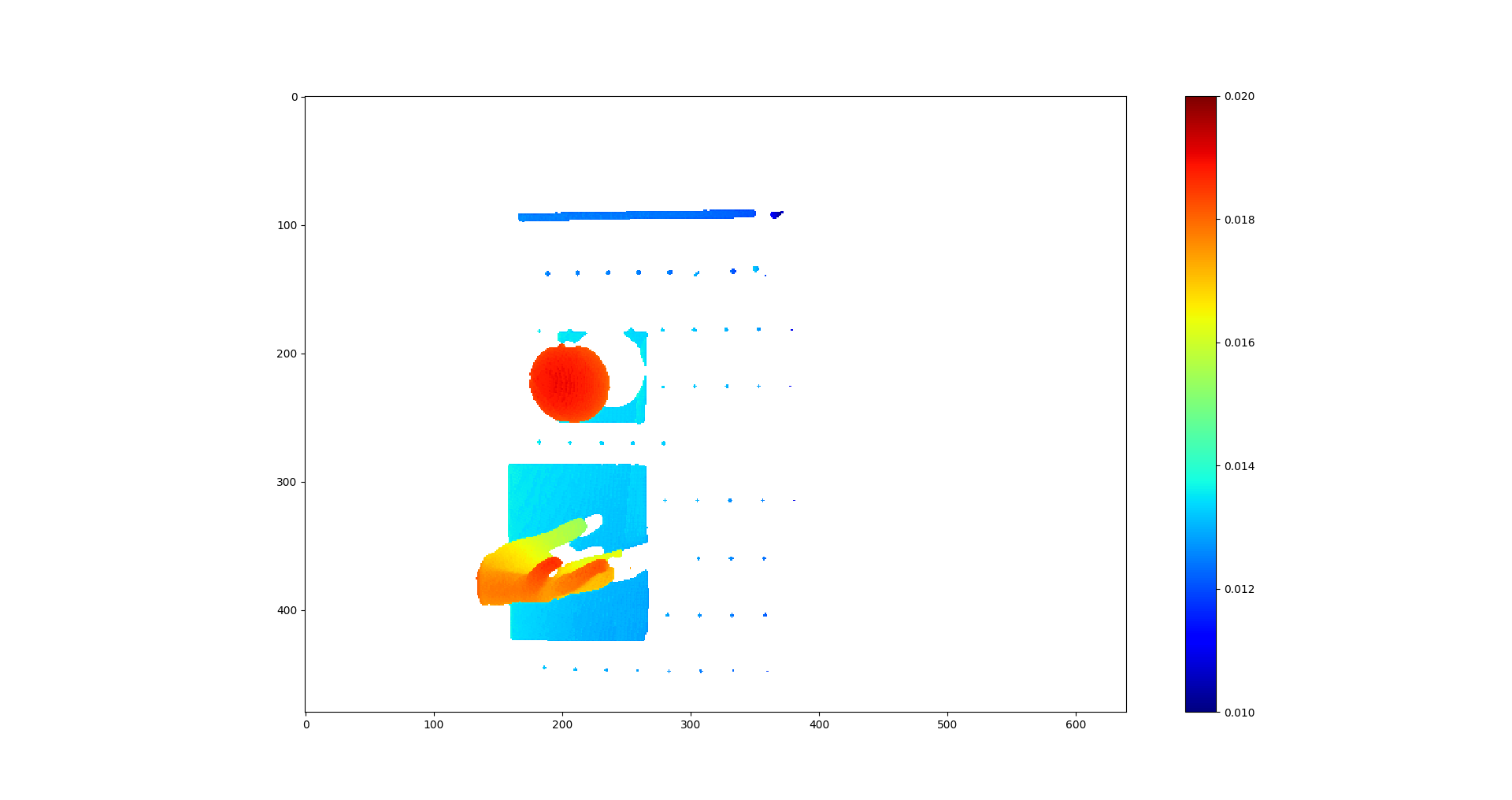}}
		\\
		
		\rotatebox{90}{\makecell{Textured plane}}
		&\includegraphics[trim={10cm 20cm 2cm 20cm},clip,width=\linewidth]{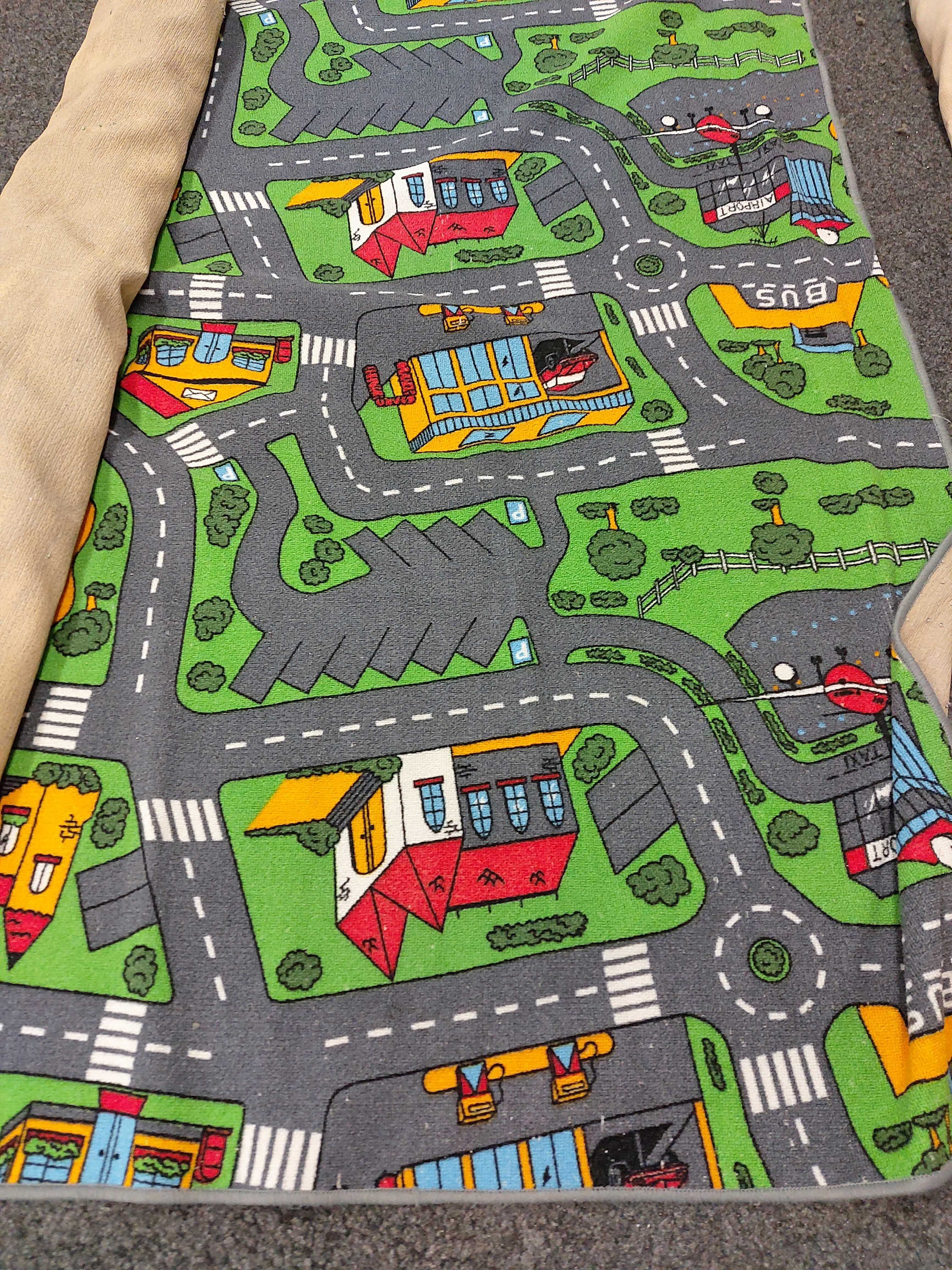}
		&\frame{\includegraphics[trim={16cm 5cm 22cm 5cm},clip,width=\linewidth]{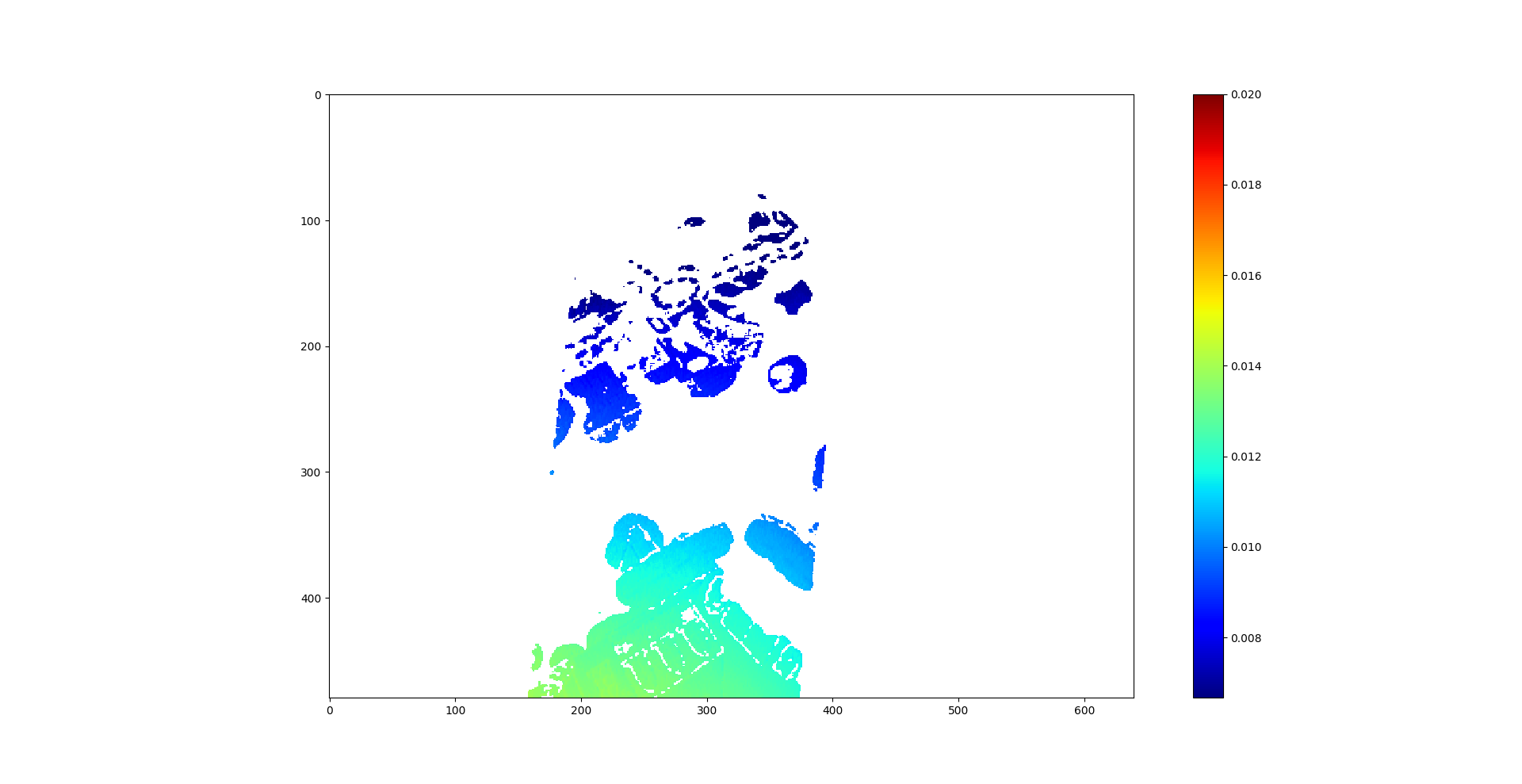}}
		&\rotatebox{90}{\makecell{Plain wall}}
		&\includegraphics[trim={2cm 0cm 2cm 5cm},clip,width=\linewidth]{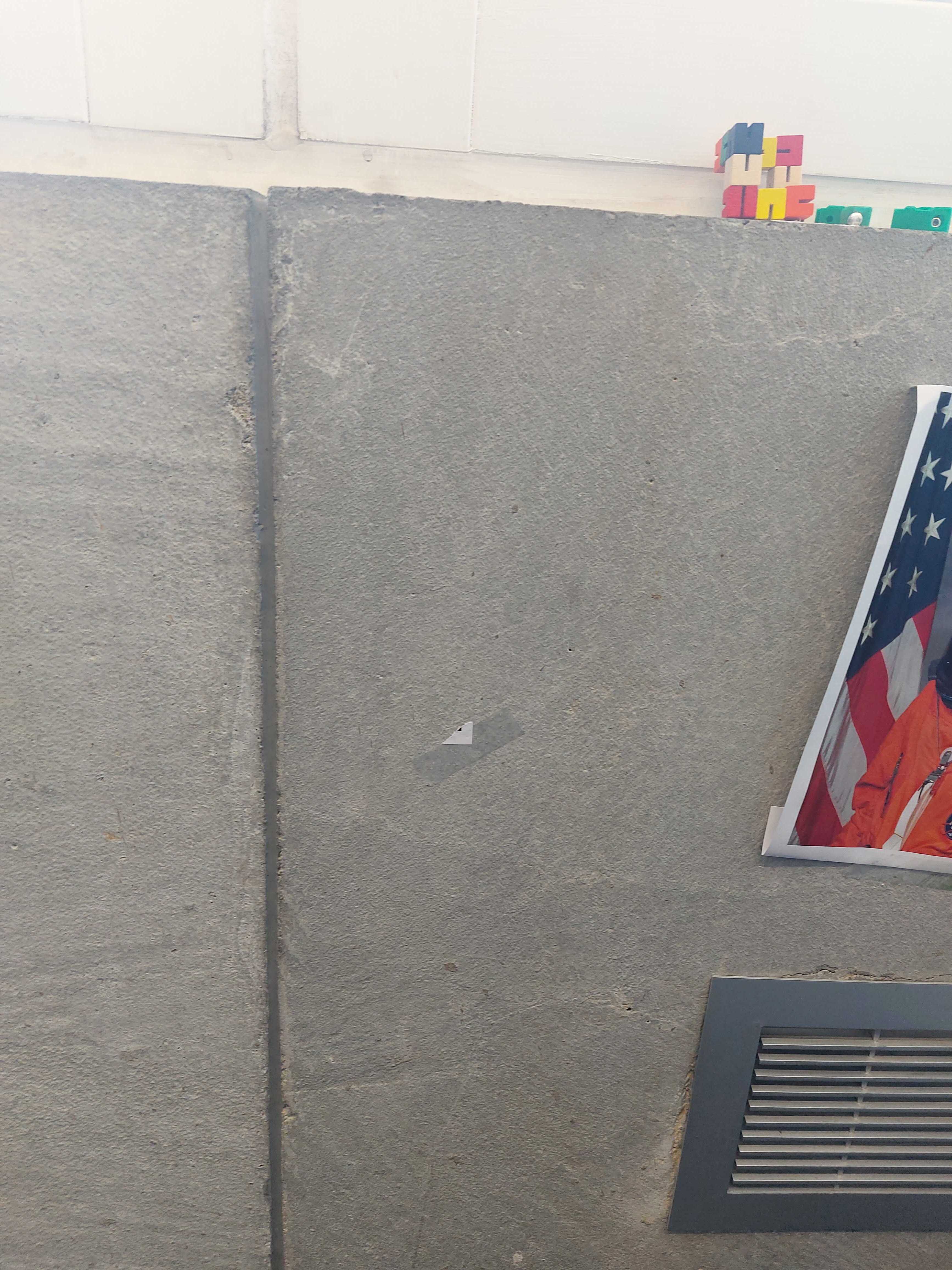}
		&\frame{\includegraphics[trim={16cm 5cm 22cm 5cm},clip,width=\linewidth]{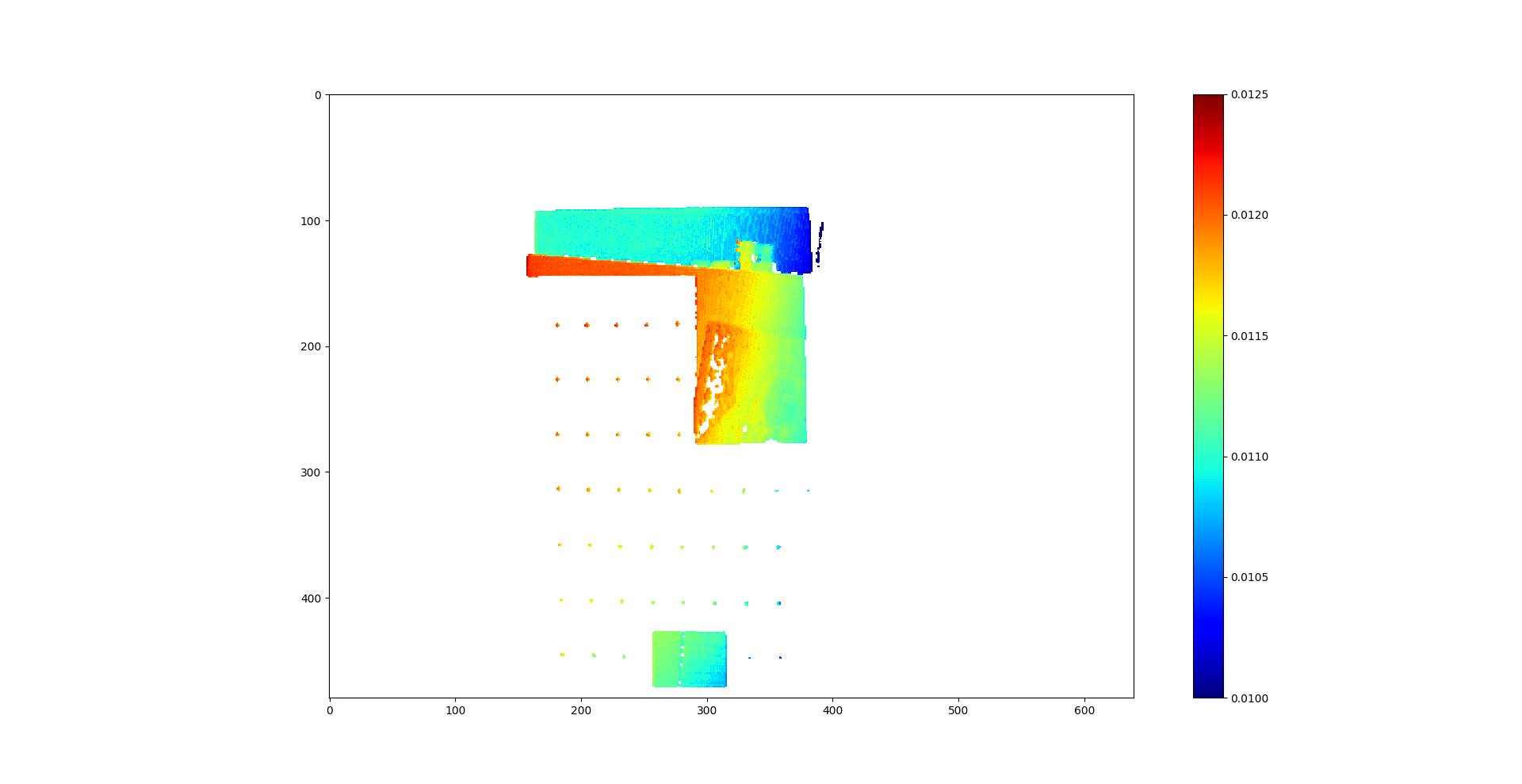}}
		\\

	\end{tabular}
	\vspace{-1ex}
	\caption{Reconstruction of dynamic scenes with real prototype.}
	\label{fig:recons}
\end{figure}

The reconstruction results can be seen in \Fig \ref{fig:recons}.
The areas without motion are sparsely sampled whereas areas with motion have dense depth.
Since we post-process the events and apply a bounding box to the active area, we can also include multiple regions of interest (RoI) as shown in the Ball sequence.
For the textureless wall, the motion of the setup generates events only in the areas where there are edges(\eg the edge of the wall and the poster edge); therefore only those areas are scanned with high spatial density.\\
\textbf{Plane fitting error}
We reconstruct a plane at a distance of $\SI{2}{\meter}$ using dense sampling, sparse sampling and event-based sampling.
The event rate (measured in $Ev/s$) significantly affects the event timestamp errors, as discussed in \Sec \ref{sec:method:noise}.
We compare the event rates and reconstruction error for three different sampling methods: dense, sparse, and event-based while scanning a plane wall.
Dense methods have a very high event rates of $6.8MEv/s$, whereas the sparse sampling has a lowest event rate of $265kEv/s$.
The event rate for event-guided depth  is data dependent and for the wall scene this varies between $600kEv/s$ and $5MEv/s$ depending on the motion.
Error is computed as the plane fitting error for the reconstruction.
The plane fitting error for  dense sampling method is $\SI{11}{\milli \meter}$, event-based sampling is $\SI{8.9}{\milli \meter}$ and sparse sampling is $\SI{7.6}{\milli \meter}$.
We observe that with sparse sampling, we achieve a lower reconstruction error because the event timestamps are accurate to produce accurate depth maps.
One the other hand, dense sampling has the highest reconstruction error due to the noise in the event timestamps present at such high event rates.
Our method strikes a balance between these two approaches and achieves a low reconstruction error and a low event-rate.
\begin{table}[t]
    \centering
    \begin{adjustbox}{max width=\columnwidth}
    \setlength{\tabcolsep}{3pt}
    \begin{tabular}{l|l|l}
    \toprule
    \textbf{Method}  & \textbf{Event rate} & \textbf{Recons. Error} \\ \midrule
    Dense & $6.8MEv/s$ &  $\SI{1.1}{\centi \meter}$ \\
    Sparse & $265kEv/s$ &  $\SI{0.76}{\centi \meter}$\\
    Event-guided &  $600kEv/s$ - $5MEv/s$ & $\SI{0.89}{\centi \meter}$ \\
    \bottomrule
    \end{tabular}
    \end{adjustbox}
    \vspace{-1ex}
    \caption{\label{tab:planerecons} Reconstruction of plane with dense, sparse and propose event-guided.}
    \vspace{-2ex}
\end{table}

\subsection{Depth completion}
\label{sec:exp:depthcompletion}
\global\long\def\figWidth{0.182\linewidth}
\global\long\def\figWidthD{0.22\linewidth}
\begin{figure*}[t]
	\centering
    \setlength{\tabcolsep}{2pt}
	\begin{tabular}{
	M{0.35cm}
	M{\figWidth}
	M{\figWidth}
	M{\figWidthD}
	M{\figWidth}
	M{\figWidthD}}
		& (a) Image & (b) Events  & (c) GT 
		& (d) Sparse samples & (e) Depth completion 
		\\

		\rotatebox{90}{\makecell{(T5\_001)}}
		&\includegraphics[width=\linewidth]{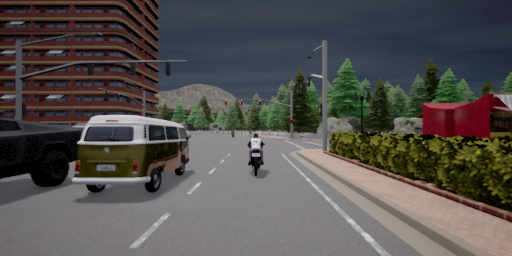}
		&\frame{\includegraphics[width=\linewidth]{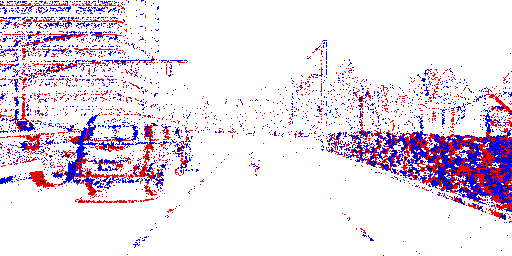}}
		&\frame{\includegraphics[trim={27cm 10cm 6cm 9cm},clip,width=\linewidth]{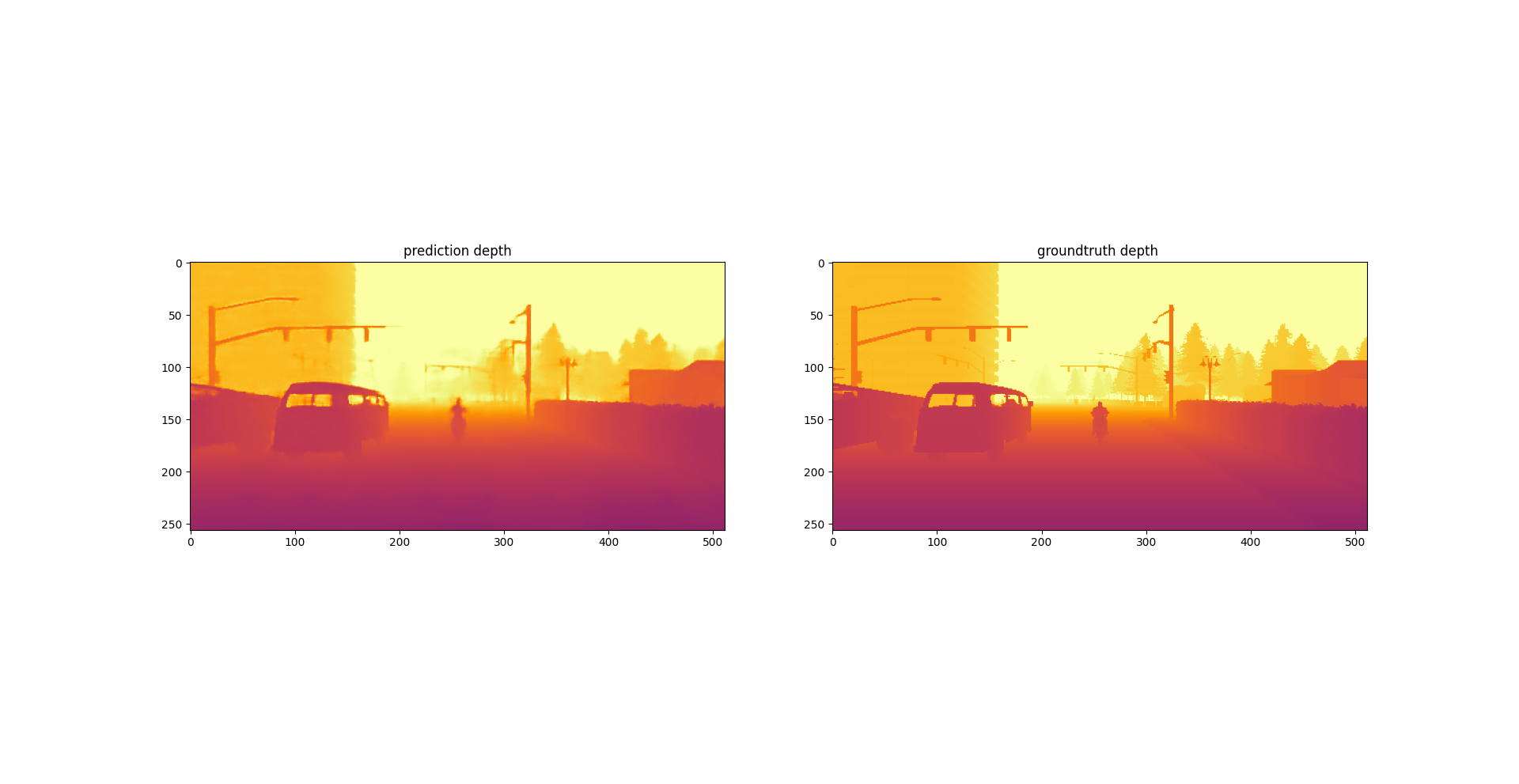}}
		&\frame{\includegraphics[trim={6cm 3cm 5cm 3cm},clip,width=\linewidth]{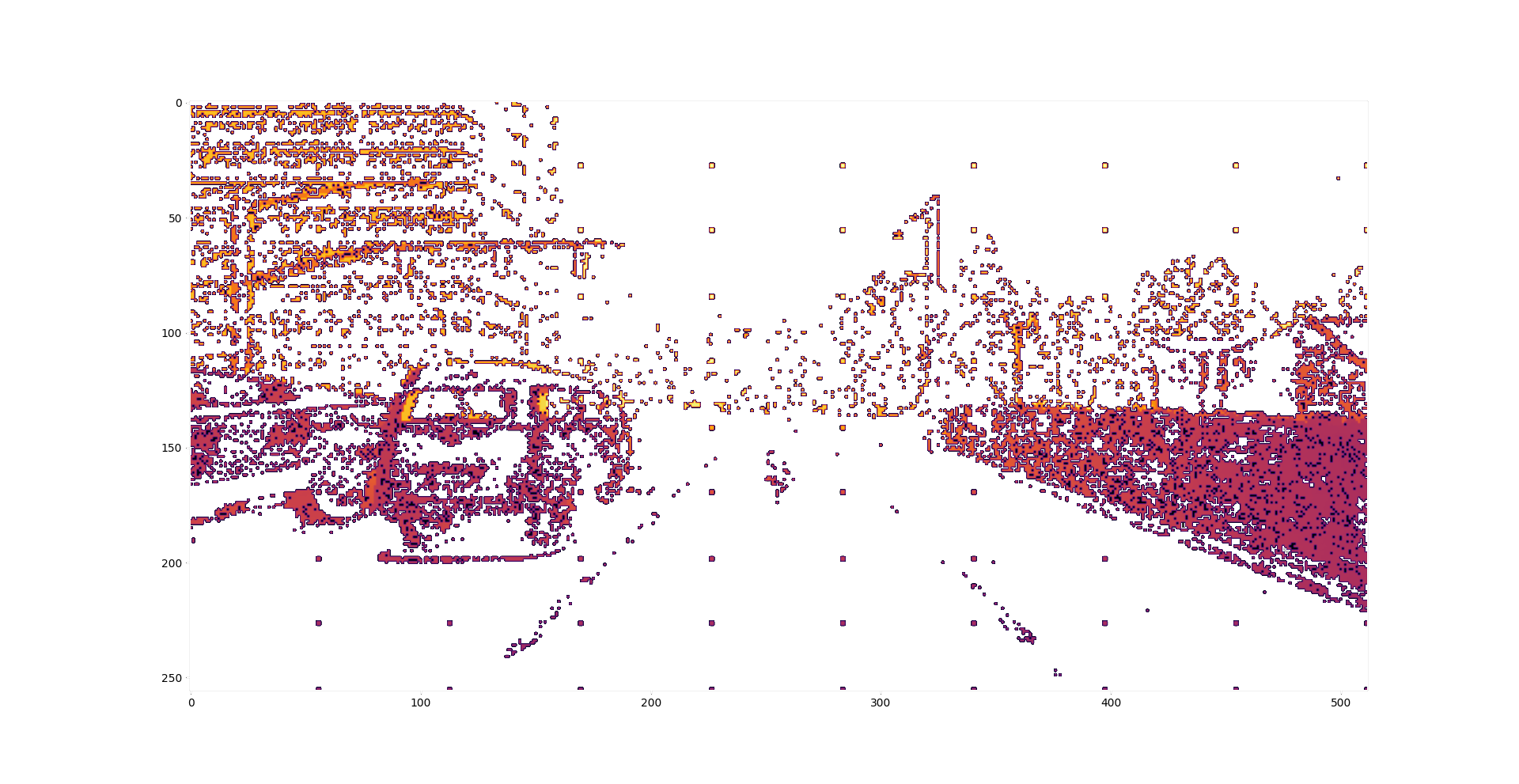}}
		&\frame{\includegraphics[trim={6.2cm 10cm 26cm 9cm},clip,width=\linewidth]{images/t5_001/prediction.png}}
		\\
		
		\rotatebox{90}{\makecell{(T5\_013)}}
		&\includegraphics[width=\linewidth]{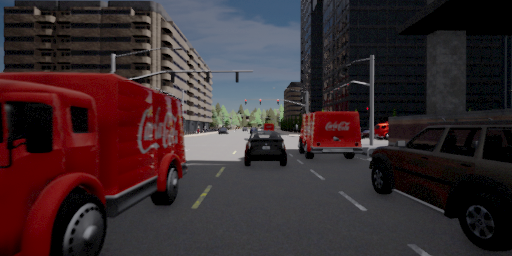}
		&\frame{\includegraphics[width=\linewidth]{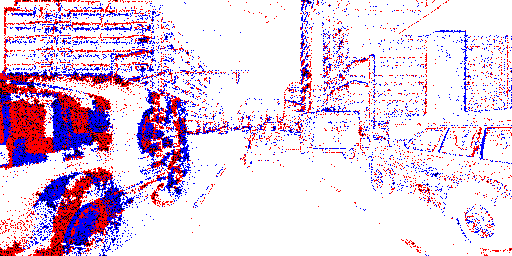}}
		&\frame{\includegraphics[trim={27cm 10cm 6cm 9cm},clip,width=\linewidth]{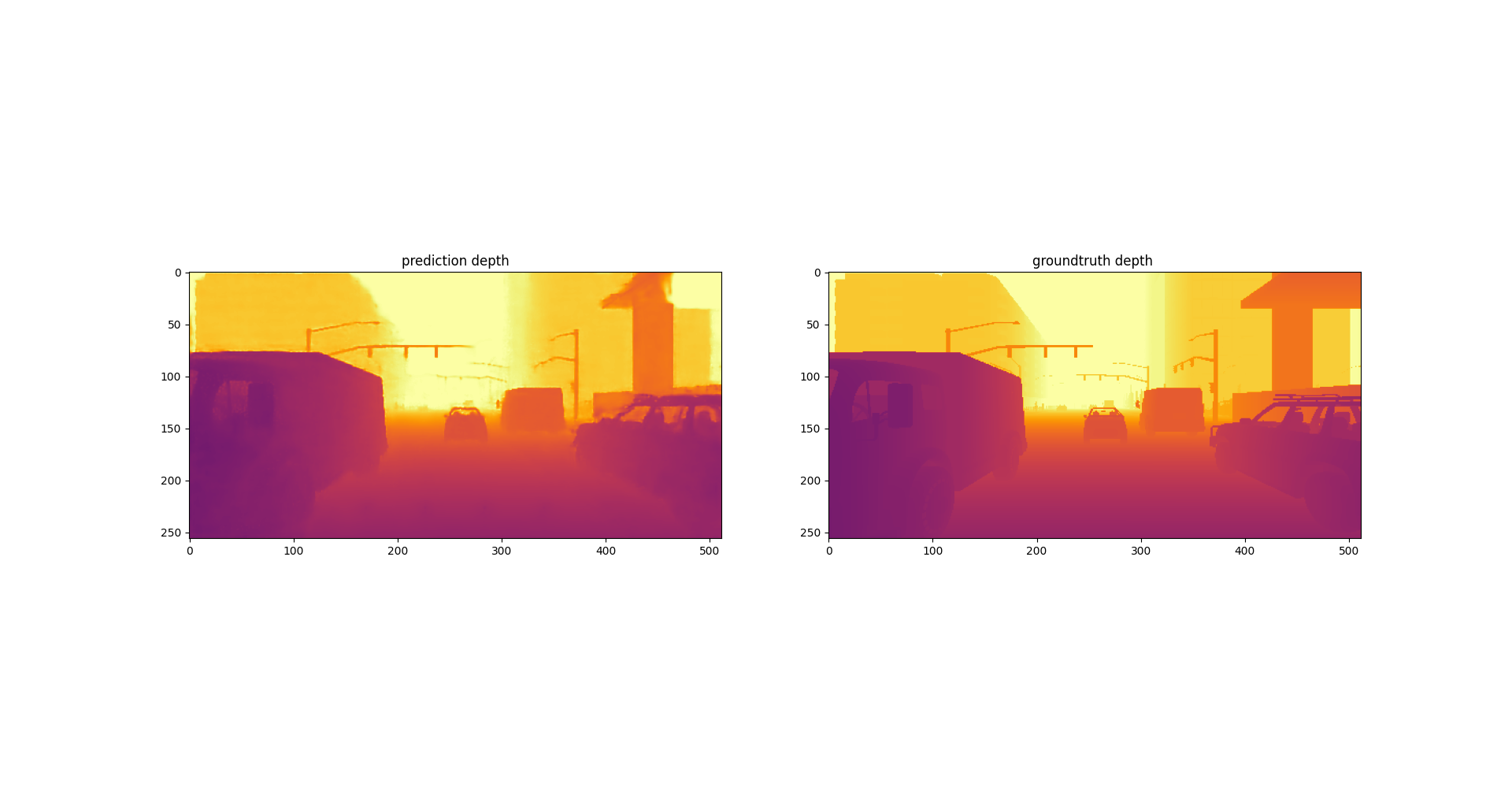}}
		&\frame{\includegraphics[trim={6cm 3cm 5cm 3cm},clip,width=\linewidth]{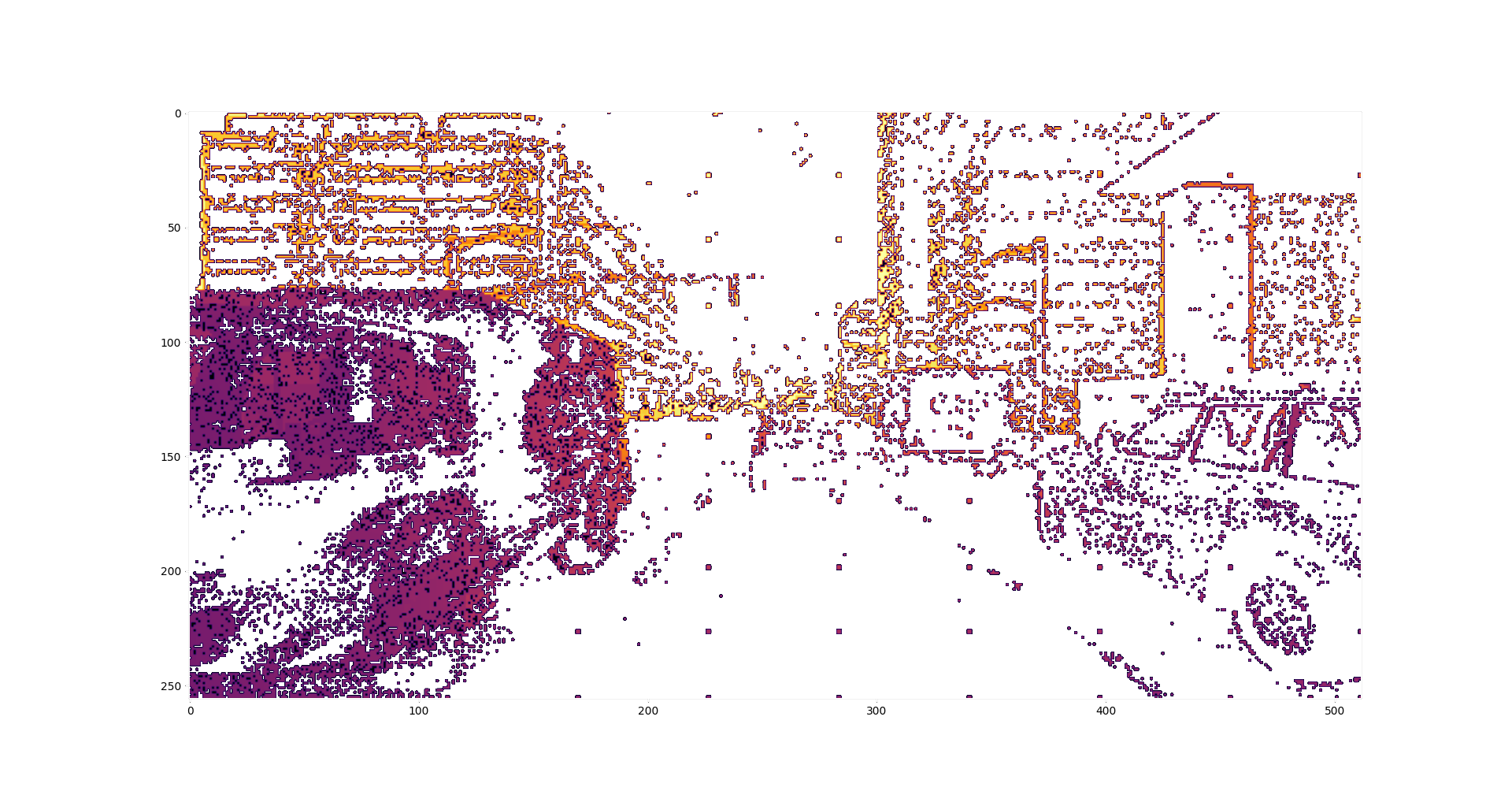}}
		&\frame{\includegraphics[trim={6.2cm 10cm 26cm 9cm},clip,width=\linewidth]{images/t5_0013_0140/prediction.png}}
		\\
		\rotatebox{90}{\makecell{(T5\_063)}}
		&\includegraphics[width=\linewidth]{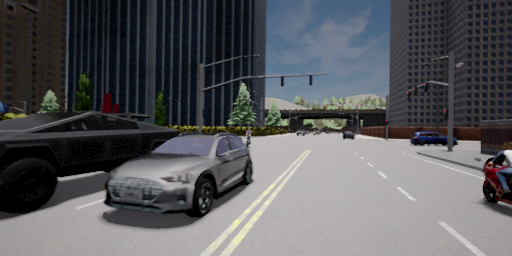}
		&\frame{\includegraphics[width=\linewidth]{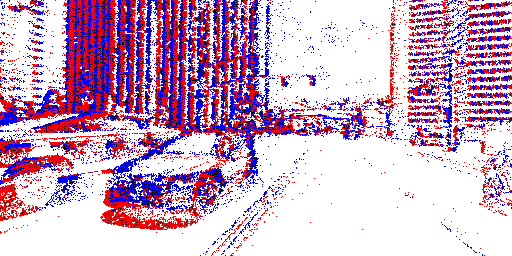}}
		&\frame{\includegraphics[trim={27cm 10cm 6cm 9cm},clip,width=\linewidth]{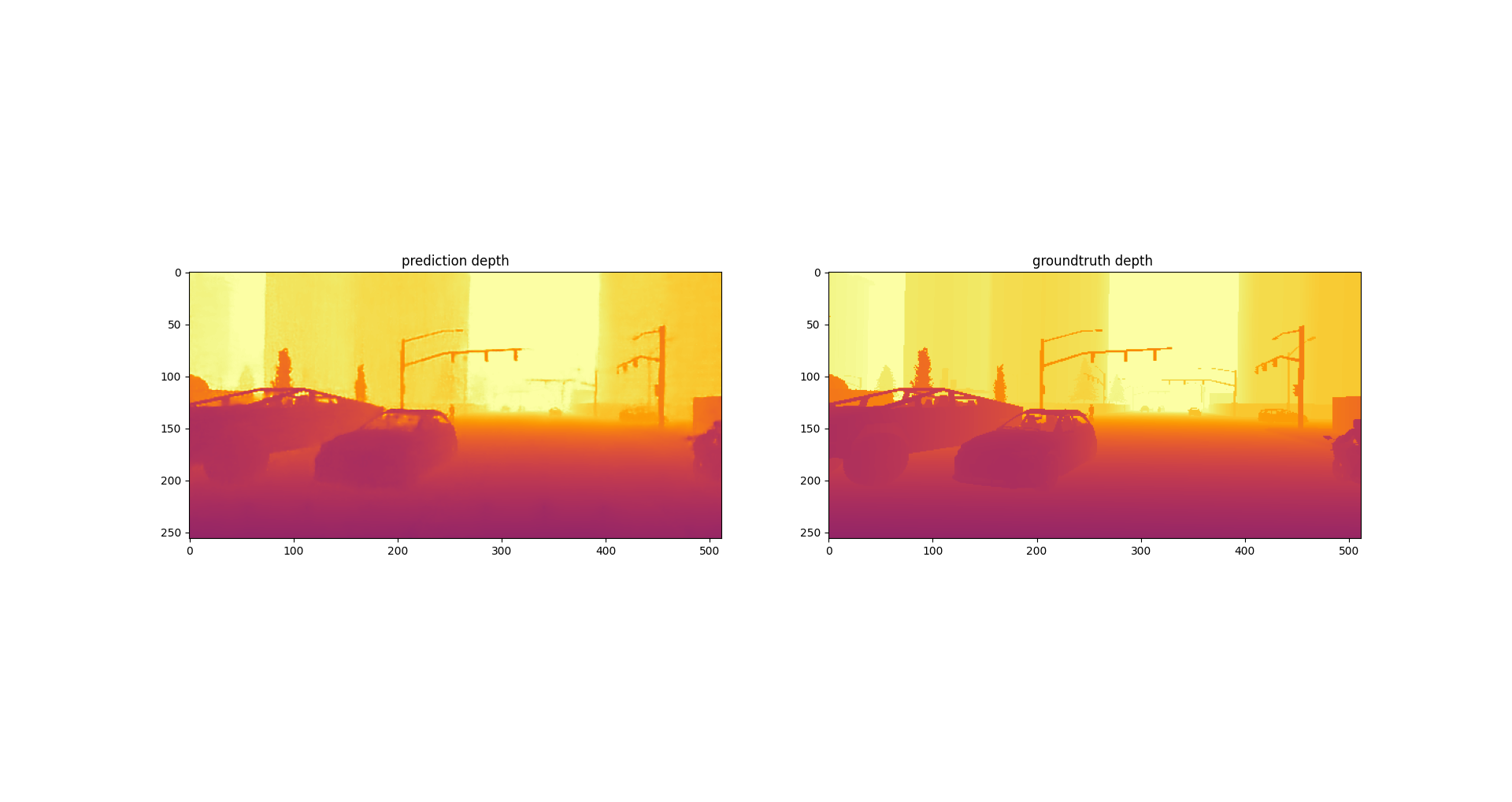}}
		&\frame{\includegraphics[trim={6cm 3cm 5cm 3cm},clip,width=\linewidth]{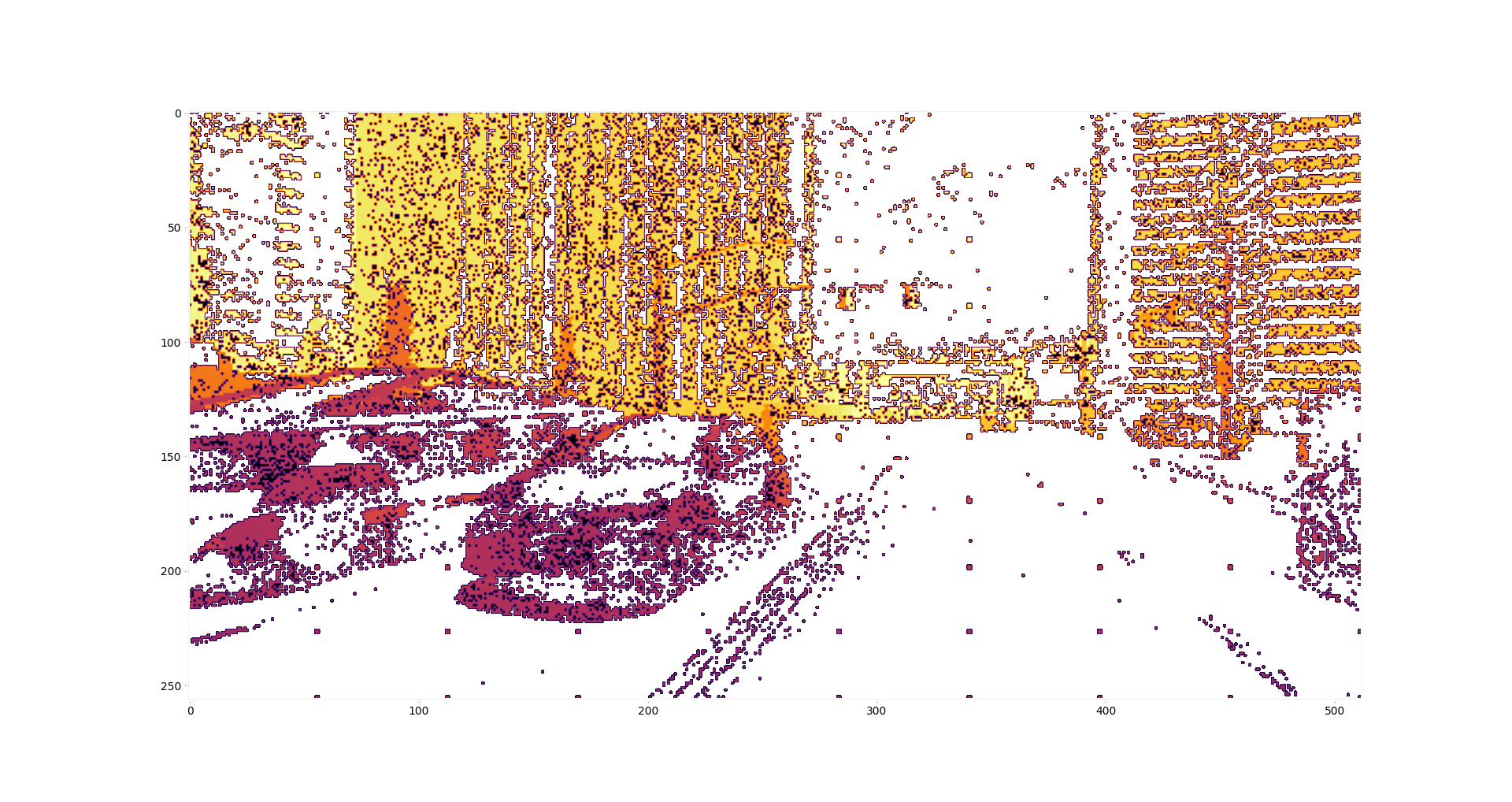}}
		&\frame{\includegraphics[trim={6.2cm 10cm 26cm 9cm},clip,width=\linewidth]{images/t5_063/prediction.png}}
		\\
		\rotatebox{90}{\makecell{(T5\_000)}}
		&\includegraphics[width=\linewidth]{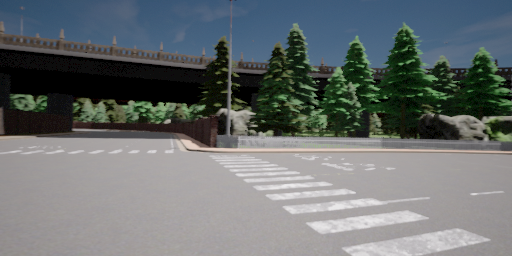}
		&\frame{\includegraphics[width=\linewidth]{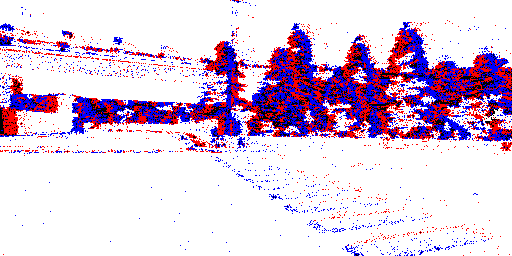}}
		&\frame{\includegraphics[trim={27cm 10cm 6cm 9cm},clip,width=\linewidth]{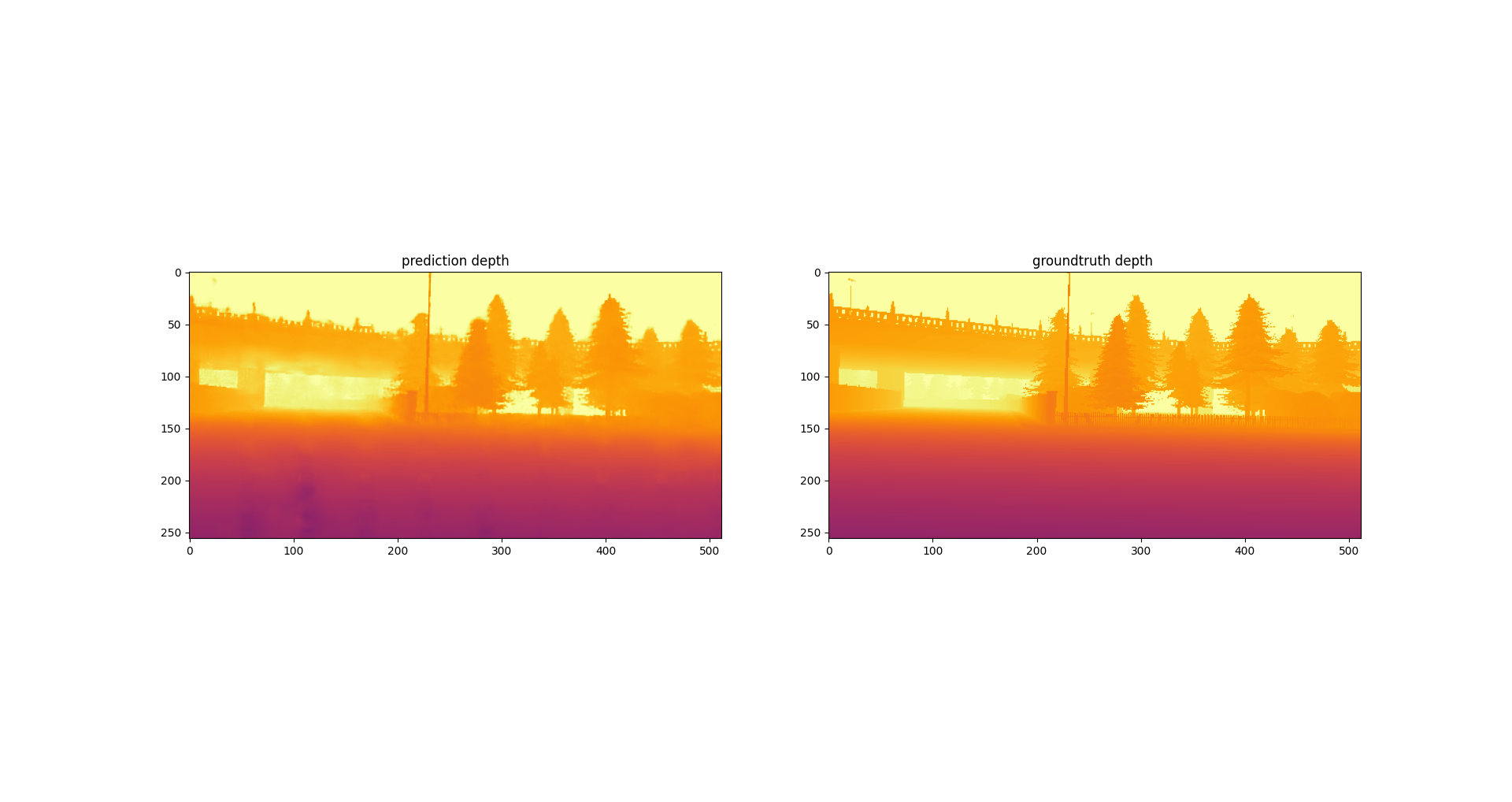}}
		&\frame{\includegraphics[trim={6cm 3cm 5cm 3cm},clip,width=\linewidth]{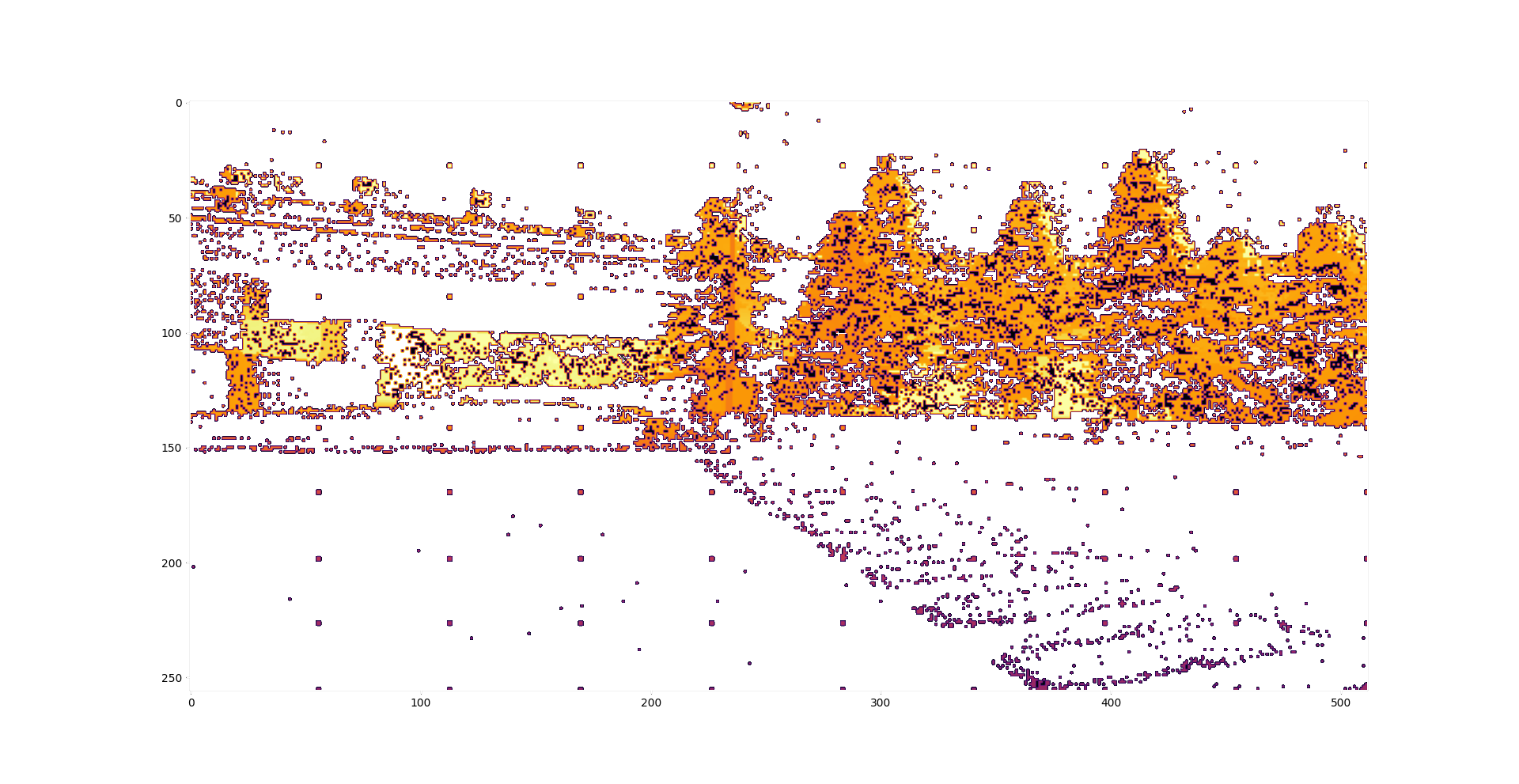}}
		&\frame{\includegraphics[trim={6.2cm 10cm 26cm 9cm},clip,width=\linewidth]{images/t5_000/prediction.png}}
	\end{tabular}
    \vspace{-1ex}
	\caption{We evaluate our method with EventScape dataset. 
	The dataset consists of (a) Images (b) Events and (c) GT Depth.
	Column (d) shows the proposed event-guided sampling and
	(e) our proposed depth completion network.
   }
	\label{fig:experim:depthcompletion}
	\vspace{-3ex}
\end{figure*}

We test our depth completion network on the test sequences of EventScape dataset.
Some qualitative results are shown in \Fig \ref{fig:experim:depthcompletion}.
As we see the events provide the sparse depth information to our network which learns to inpaint the depth.
Since events correspond to moving edges, the sparse depth measurements also correspond to the edges and therefore the output of our depth prediction network is sharp around the edges.
However in areas which are textureless and planar like $T_{013}$ no events are generated and therefore no sparse measurements are generated leading to wrong depth estimates in such areas.
The proposed depth completion method can recover full frame depth from sparse event-guided depth input with an accuracy of $\SI{1.32}{\centi \meter}$.

\section{Conclusion}
\label{sec:conclusion}
We present a novel bio-inspired event-guided depth estimation method.
Inspired from the human ability to focus with higher resolution on areas of interest and decrease the resolution for background, we propose event-camera based depth sampling as events naturally encode relative motion of the scene.
We show that in natural scenes like autonomous driving and indoor environments moving edges correspond to less than 10\% of the scene.
Thus our setup requires the sensor to scan only 10\% of the scene.
This would equate to almost 90\% less power consumption by the illumination source.%
\mm{We demonstrate the capabilities of this system by building a prototype and qualitatively demonstrate the depth reconstruction in office scenarios.}

{\small
\bibliographystyle{ieeetr_fullname} %
\bibliography{main}
}

\end{document}